\documentclass{article}

\usepackage{microtype}
\usepackage{graphicx}
\usepackage{subfigure}
\usepackage{booktabs} 
\usepackage{multirow}

\usepackage{hyperref}

\usepackage[accepted]{icml2024}

\usepackage{amsmath}
\usepackage{amssymb}
\usepackage{mathtools}
\usepackage{amsthm}
\usepackage{threeparttable}

\usepackage[capitalize,noabbrev]{cleveref}

\theoremstyle{plain}

\theoremstyle{definition}

\theoremstyle{remark}

\newcommand\T{\rule{0pt}{2.4ex}}       

\newcommand{\icmlIntern}{\textsuperscript{*} Work done as an intern in PolyU.}
\usepackage[textsize=tiny]{todonotes}

\icmltitlerunning{Leddam: Learnable Decomposition with Inter-Series Dependencies and Intra-Series Variations Modeling}

\begin{document}

\twocolumn[
\icmltitle{Revitalizing Multivariate Time Series Forecasting: Learnable Decomposition with Inter-Series Dependencies and Intra-Series Variations Modeling}


\icmlsetsymbol{intern}{*}

\begin{icmlauthorlist}
\icmlauthor{Guoqi Yu}{polyu,yyy,intern}
\icmlauthor{Jing Zou}{polyun}
\icmlauthor{Xiaowei Hu}{sh}
\icmlauthor{Angelica I. Aviles-Rivero}{cam}
\icmlauthor{Jing Qin}{polyun}
\icmlauthor{Shujun Wang}{polyu}
\end{icmlauthorlist}
\icmlaffiliation{yyy}{University of Electronic Science and Technology of China, Sichuan, China}
\icmlaffiliation{sh}{Shanghai Artificial Intelligence Laboratory, Shanghai, China}
\icmlaffiliation{polyu}{Department of Biomedical Engineering, The Hong Kong Polytechnic University, Hong Kong SAR, China}
\icmlaffiliation{polyun}{School of Nursing, The Hong Kong Polytechnic University, Hong Kong SAR, China}
\icmlaffiliation{cam}{DAMTP, University of Cambridge, Cambridge, UK}
\icmlcorrespondingauthor{Shujun Wang}{shu-jun.wang@polyu.edu.hk}
\icmlkeywords{Multivariate Time Series Prediction, Learnable Decomposition Strategy, and Time Series Forecasting}

\vskip 0.3in
]
%
\printAffiliationsAndNotice{\icmlIntern}

\begin{abstract}
Predicting multivariate time series is crucial, demanding precise modeling of intricate patterns, including inter-series dependencies and intra-series variations. 
Distinctive trend characteristics in each time series pose challenges, and existing methods, relying on basic moving average kernels, may struggle with the non-linear structure and complex trends in real-world data. 
Given that, we introduce a learnable decomposition strategy to capture dynamic trend information more reasonably.
Additionally, we propose a dual attention module tailored to capture inter-series dependencies and intra-series variations simultaneously for better time series forecasting, which is implemented by channel-wise self-attention and autoregressive self-attention.
To evaluate the effectiveness of our method, we conducted experiments across eight open-source datasets and compared it with the state-of-the-art methods. 
Through the comparison results, our \textbf{Leddam} (\textbf{LE}arnable \textbf{D}ecomposition and \textbf{D}ual \textbf{A}ttention \textbf{M}odule) not only demonstrates significant advancements in predictive performance but also the proposed decomposition strategy can be plugged into other methods with a large performance-boosting, from 11.87\% to 48.56\% MSE error degradation. Code is available at this link: \href{https://github.com/Levi-Ackman/Leddam}{https://github.com/Levi-Ackman/Leddam}.
\end{abstract}    
\section{Introduction}

\begin{figure*}
  \centering
  \includegraphics[width=\linewidth]{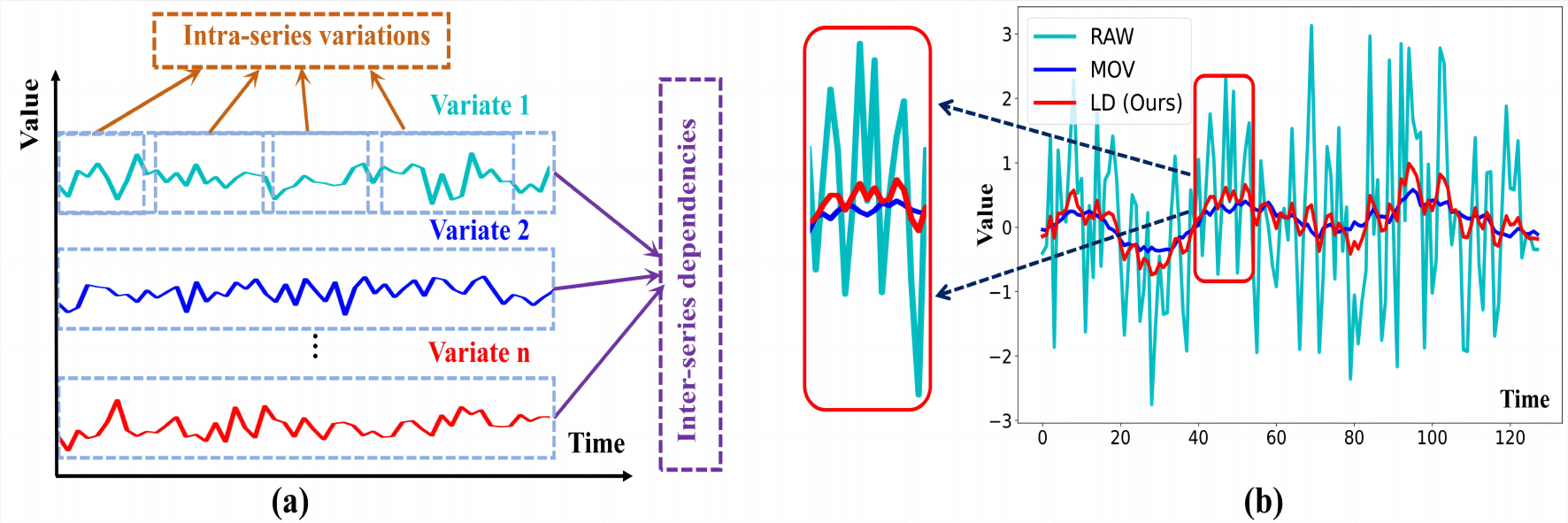}
  \caption{(a) Demonstration of inter-series dependencies and intra-series variations. (b) Visualization of different decomposition schemes in Electricity data. RAW means the raw time series. MOV means moving average kernel, and LD means our learnable decomposition module.}
  \label{fig:intro}
\end{figure*}

The rising demands in diverse real-world domains have generated an urgent requirement for precise multivariate time series forecasting methodologies, as demonstrated in fields like energy management~\citep{dong2023simmtm,wu2024stanhop,yi2023fouriergnn}, weather forecasting~\citep{anonymous2024moderntcn,anonymous2024timemixer,anonymous2024fits}, disease control~\citep{yi2023fmlp,Liu2023koopa,Zhou2022film,Ni2023basiformer},  and traffic planning~\citep{Rangapuram2018dpstate,Zhao2017traflstm,Shao2022trafgnn}.
The foundation of precise forecasting models lies in effectively identifying and modeling intricate patterns embedded in multivariate time series. Two primary patterns (Figure~\ref{fig:intro}(a)) emerge as inter-series dependencies and intra-series variations~\citep{zhang2023crossformer}.
The former delineates the intricate interplay and correlations among distinct variables, while the latter encapsulates both enduring and ephemeral fluctuations within each specific time series.

However, the time series of each constituent variable in multivariate time series data often displays distinctive variations in its trends. Such discrepancies inherent in raw time series may complicate the modeling of inter-series dependencies~\citep{LiuiTransformer}.
Furthermore, time series in the real world are continually susceptible to distributional shifts induced by the evolution of their trends—a distinctive attribute that adds complexity to modeling the dynamic intra-series variation patterns within the sequences~\citep{Taylor_2018_distribution,liu2022non}.
Therefore, a robust forecasting method should be capable of addressing the following two challenges: (1) How to precisely unravel patterns within raw time series under the interference of trend components. (2) How to efficiently model inter-series dependencies and intra-series variations.

Certain studies~\citep{Wu2021autoformer,Zhou2022fedformer,wang2023micn,zeng2023dlinear} aim to address the first challenge by conducting trend-seasonal decomposition of the original time series using Moving Average kernel (MOV).
However, such an untrainable procedure along with moving average kernels leads to a lack of robustness.
Moreover, the uniform assignment of weights to each data point within the sliding window may impede their ability to discern specific patterns. 
Such limitation becomes apparent when dealing with intricate time series data (RAW data), particularly those with non-linear structures or significant noise levels, as illustrated in Figure~\ref{fig:intro}(b) (RAW VS. MOV).
Therefore, it is necessary to develop a learnable decomposition method to revitalize multivariate time series forecasting tasks.

For the second challenge of modeling inter-series dependencies and intra-series variations in time series forecasting, recent efforts have turned to Transformer architectures renowned for their robust pairwise dependency delineation and multi-level representation extraction within sequences~\citep{Vaswani2017transformer}.
iTransformer~\citep{LiuiTransformer} effectively handles inter-series dependencies through `Channel-wise self-attention', embedding entire time series into a token but lacks explicit learning of intra-series variations.
Other works directly employ pair-wise attention mechanisms for intra-series variations~\citep{Liu2021pyraformer,Zhou2020informer,Li2019logtrans,liu2022non}. However, they improperly use permutation-invariant attention mechanisms on the temporal dimension~\citep{zeng2023dlinear}.
Alternatively, some approaches shift to partitioning time series into patches and applying self-attention modeling on these patches~\citep{Nie2022patchtst,zhang2023crossformer}. Yet, such methods inherently lead to information loss as patches encapsulate only a portion of the original sequence. Moreover, the optimal patch length is also hard to determine.
Therefore, we aim to deal with it by generating features suitable for modeling intra-series variations while maximizing information preservation to avoid the above disadvantages.

In this paper, we aim to revitalize multivariate time series forecasting with \textbf{Leddam} (\textbf{LE}arnable \textbf{D}ecomposition and \textbf{D}ual \textbf{A}ttention \textbf{M}odule).
Specifically, we first introduce a trainable decomposition module
to decompose the original time series data into more reasonable Trend and Seasonal parts. 
This allows the kernel to prioritize the present data point and adapt to non-linear structures or noise in raw time series, capturing dynamic trend information effectively (see LD in Figure \ref{fig:intro}(b)).
Secondly, we design a `Dual Attention Module', where 1) channel-wise self-attention to capture inter-series dependencies; 2) an enhanced methodology involving an auto-regressive process and attention mechanism on generated tokens to model intra-series variations.
Our \textbf{Leddam} aims to provide a more robust and comprehensive solution to time series forecasting challenges.
The primary contributions are summarized as follows. 
\vspace{-0.2cm}
\begin{itemize}
\item We propose the incorporation of a learnable convolution kernel initialized with a Gaussian distribution to enhance time series decomposition.
\vspace{-0.2cm}
\item We devise a `Dual Attention Module' that adeptly captures both inter-series dependencies and intra-series variations concurrently.
\vspace{-0.2cm}
\item We validate our \textbf{Leddam} by showing that not only demonstrates significant advancements in predictive performance but also the proposed decomposition strategy can be plugged into other methods with a large performance boosting, from \textbf{11.87\%} to \textbf{48.56\%} MSE error degradation.
\end{itemize}

\section{Related work}

\textbf{Time Series Data Decomposition.} Due to the capacity of the moving average kernel to smooth out short-term fluctuations or noise in the time series, Autoformer~\citep{Wu2021autoformer} initially proposed employing the moving average kernel for extracting the trend part of the time series. Later works, including MICN~\citep{wang2023micn}, FEDformer~\citep{Zhou2022fedformer}, DLinear~\citep{zeng2023dlinear}, etc., have predominantly adhered to their methodology. However, a rudimentary averaging kernel may inadequately capture precise trends in time series characterized by more intricate patterns than simple linear relationships. It applies uniform weighting to all data points within the window size, which may not be suitable for capturing non-linear or non-stationary trends present in the data.

\textbf{Intra-series Variations Modeling.} It is important to comprehend the temporal variations inherent within the time series for a precise time series forecasting model. Due to the intrinsic limitations of point-to-point attention mechanisms, such as those employed in models like Informer~\citep{Zhou2020informer}, Reformer~\citep{Kitaev2020reformer}, and Pyraformer~\citep{Liu2021pyraformer}, the resultant attention maps are prone to suboptimality. This arises from the fact that individual points in a time series, in contrast to words~\citep{Vaswani2017transformer} or image patches~\citep{Dosovitskiy2020vit}, lack explicit semantic information. Subsequent endeavours involve partitioning the primary time series into a series of patches~\citep{zhang2023crossformer,Nie2022patchtst}, followed by applying self-attention mechanisms across these patches to model temporal variations. However, segmenting time series into patches inevitably introduces information loss.

\textbf{Inter-series Dependencies Modeling.} Inter-series dependencies constitute a pivotal attribute that distinguishes multivariate time series from their univariate counterparts. It has come to our attention that the majority of transformer-based methodologies opt to treat values from different variables at the same time step or from distinct channels as tokens~\cite{Zhou2020informer,Kitaev2020reformer,Liu2021pyraformer,liu2022non} to model inter-series dependencies. Such strategies may result in attention maps that lack meaningful information, consequently impeding the effective and accurate modeling of the information we seek~\citep{LiuiTransformer}. Certain endeavours have sought to adopt channel-independent designs, aiming to mitigate the reduction in predictive accuracy induced by this operation, as exemplified by PatchTST~\citep{Nie2022patchtst} and DLinear~\citep{zeng2023dlinear}. Channel Independence (CI) regarding variates of time series independently and adopting the shared backbone, have gained consistently increasing popularity in forecasting with performance promotions as an architecture-free method. Recent works~\citep{Han2023ci1,Li2023ci2} found that while Channel Dependence (CD) benefits from a higher capacity ideally, CI can greatly boost the performance because of sample scarcity, since most of the current forecasting benchmarks are not large enough. However, neglecting the interdependencies among variables may lead to suboptimal outcomes~\citep{LiuiTransformer}. Crossformer~\citep{zhang2023crossformer} strives to partition time series into patches and subsequently engage in the learning of inter-series dependencies on these patches, a process which, as argued previously, may engender information loss. iTranformer~\citep{LiuiTransformer} achieves more precise modeling of relationships among variables of multivariate time series by embedding the entire time series of a variate into a token, thereby avoiding information loss.

\section{Methodology}
\begin{figure}
  \centering
  \includegraphics[width=0.85\linewidth]{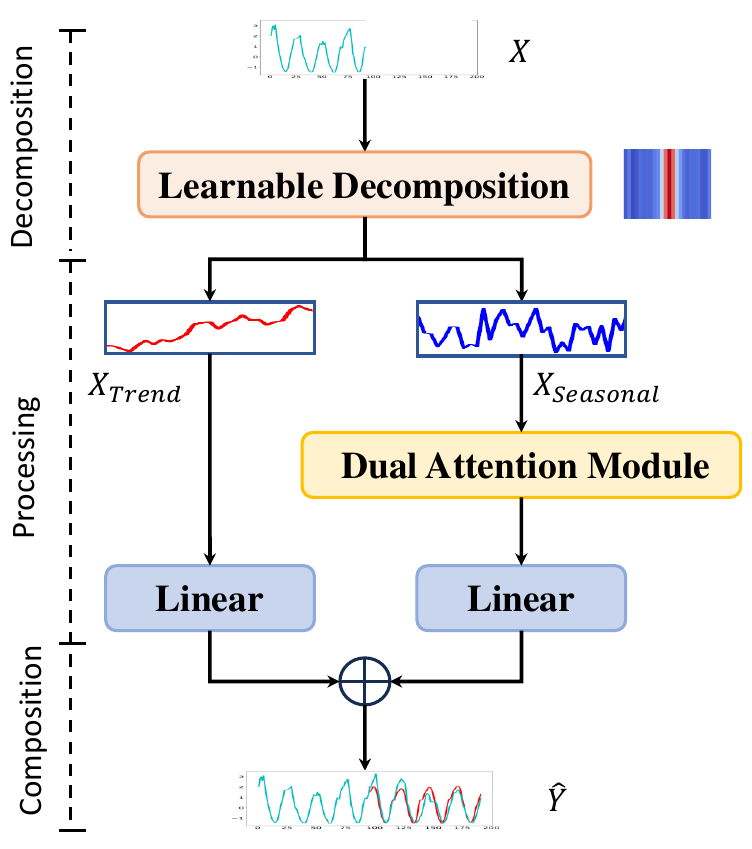}
  \caption{Overall structure of proposed \textbf{Leddam}. We start by embedding the time series and incorporating positional encoding. Then, the time series is decomposed into its trend and seasonal parts, each addressed through distinct methodologies. Finally, the processed outcomes of these two components are aggregated to obtain the ultimate predictive result. }
  \label{fig:DEFT}
\end{figure}

In this section, we elucidate the overall architecture of \textbf{Leddam}, which is depicted in Figure \ref{fig:DEFT}.
%
%
We will first define the problem and then describe the proposed Learnable Convolutional Decomposition strategy and Dual Attention Module.

\subsection{Problem Definition}

Given a multivariate time series input $X \in \mathbb{R}^{N \times T}$, time series forecasting tasks are designed to predict its future $F$ time steps $\hat{Y}\in \mathbb{R}^{N \times F}$, where $N$ is the number of variates or channels, and $T$ represents the look-back window length.
We aim to make $\hat{Y}$ closely approximate $Y \in \mathbb{R}^{N\times F}$, which represents the ground truth.

\subsection{Learnable Decomposition Module}

We employ a superior learnable 1D convolutional decomposition kernel instead of a moving average kernel to comprehensively encapsulate the nuanced temporal variations in the time series. 

\begin{figure*}
  \centering
  \includegraphics[width=0.85\linewidth]{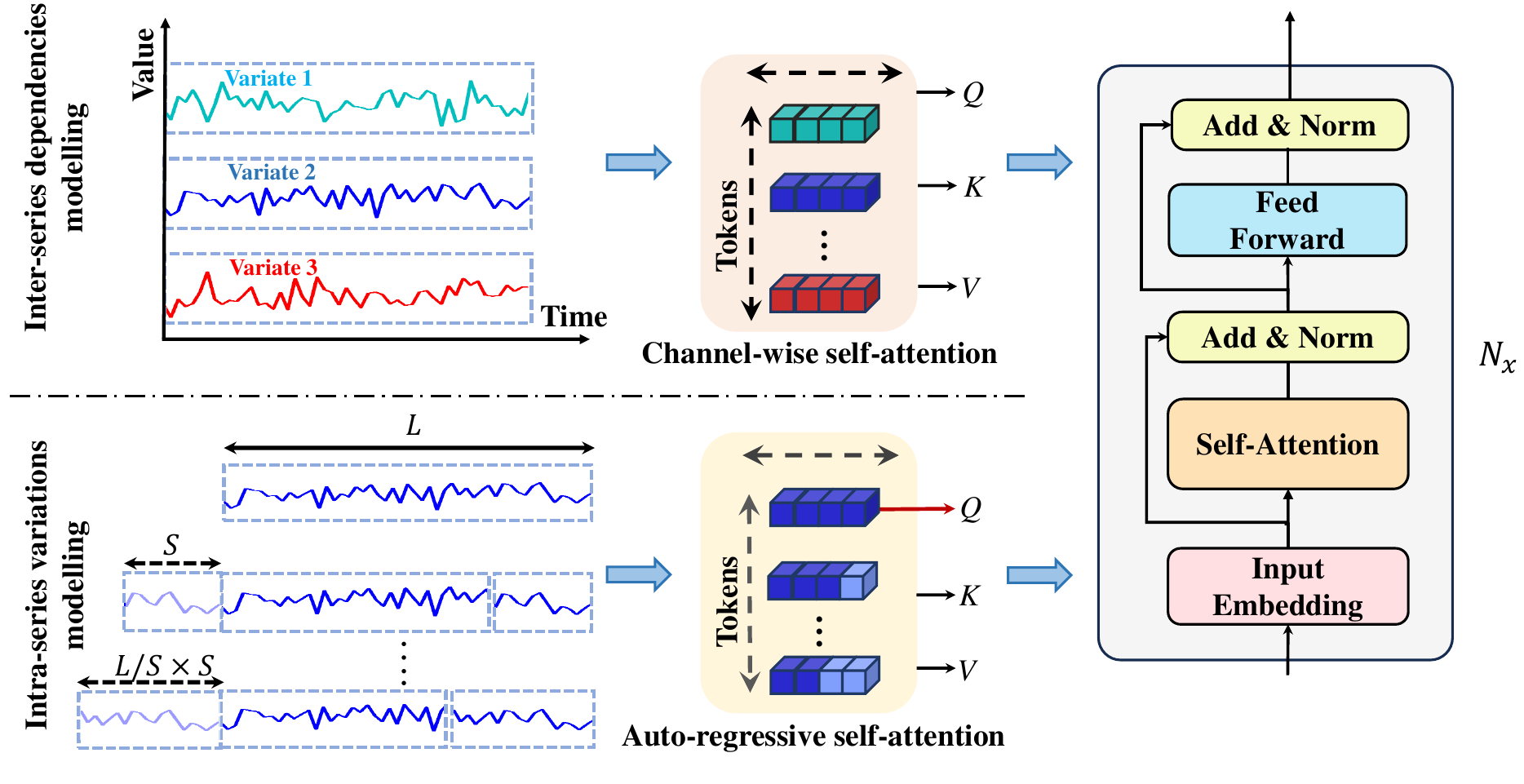}
  \caption{General process of `Dual Attention Module' to deal with Inter-series dependencies and Intra-series variations, respectively. `Channel-wise self-attention' embeds the whole series of a channel to generate `Whole Series Embedding', and transformer encoders are employed to model Inter-series dependencies. `Auto-regressive self-attention' generates `Auto-regressive Embedding' and still utilizes transformer encoders to model Intra-series variations}
  \label{fig:Two-embedding}
\end{figure*}

\textbf{Projection and position embedding.} Following iTransformer~\citep{LiuiTransformer}, we first map time series data $X \in \mathbb{R}^{N \times T}$ from the original space to a new space, subsequently incorporating positional encoding $Pos\in \mathbb{R}^{N \times D}$ as $X_{embed} \in \mathbb{R}^{N \times D}$ following $X_{embed} = (XW  + b)+Pos,$
with weights $W \in \mathbb{R}^{T \times D}, b \in \mathbb{R}^{1 \times D}$,
where $D$ is the dimension of the layer.

\textbf{Learnable 1D convolutional decomposition kernel.}
To realize the learnable convolutional decomposition, we need to define the convolutional decomposition kernel first.
Specifically, we pre-define a stride of $S = 1$ and a kernel size of $K = 25$ experimentally. Regarding its weight, initialization is performed utilizing a Gaussian distribution. We assume its weight is $\omega \in \mathbb{R}^{1 \times 1 \times K}$, and a hyperparameter $\sigma \in \mathbb{R}$. Here we set $\sigma = 1.0$. Then, we have
\begin{align}
    U \in \mathbb{R}^{1 \times 1 \times K}& ,\notag\\
    U[0, 0, i] = \exp\left(-\frac{(i - K/2)^2}{2\sigma^2}\right)&,\ i = 1, 2, \ldots, K, \notag\\
    \omega = \text{Softmax}(U, \text{dim}&=-1).
\end{align}
Therefore, this initialization results in the central position of the convolutional kernel having the maximum weight, while the edge positions of the kernel have relatively smaller weights. This is typically beneficial for convolutional layers to be more sensitive to the central position when recognizing specific features. Given a multivariate time series input $X_{embed} \in \mathbb{R}^{N \times D}$, where $N$ represents the dimensionality of channels, corresponding to the number of variables in the time series, and $D$ is the dimensionality of the embedding after positional and temporal encoding, to maintain the equivalence of sequence lengths before and after convolution, padding is employed using terminal values. So we get $X_{padded} \in \mathbb{R}^{N \times (D+K-1)}$, we split it into $N$ individual time series $x_{i} \in \mathbb{R}^{1 \times (D+K-1)}, \quad i = 1, 2, \ldots, N$. Subsequently, for each $x_{i}$, we apply a learnable 1D convolutional kernel with shared weights to extract its trend component denoted as $\hat{x_{i}}\in \mathbb{R}^{1 \times D}$. Subsequently, the convolutional outputs of all $\hat{x_{i}}$ are concatenated to form the resultant matrix $X_{Trend} \in \mathbb{R}^{N \times D}$. We get the seasonal part $X_{Seasonal} \in \mathbb{R}^{N \times D}$ by $X_{Seasonal} = X_{embed}-X_{Trend}$. The whole process can be summarized as 
\begin{align}
    X_{Trend}  &= LD(Padding(X_{embed})),\notag\\
    X_{Seasonal} &= X_{embed}-X_{Trend}.
\end{align}
\vspace{-0.5cm}

\textbf{Trend part.} Given the smoother and more predictable
nature of the trend part, we employ a simple MLP 
for projection to derive the trend part's output
akin to~\citep{zeng2023dlinear,wang2023micn}, which reads:
\begin{align}
    W \in \mathbb{R}^{D \times F},b \in \mathbb{R}^{1 \times F}, \notag\\
    X_{T_{out}} = X_{Trend}W  + b.
\end{align}

\textbf{Seasonal part.} 
Considering the suitability of the seasonal component for modeling inter-series dependencies and intra-series variations,
we transform $X_{Seasonal} \in \mathbb{R}^{N \times D}$ into two distinct embeddings: `Whole Series Embedding' and `Auto-regressive Embedding'. This facilitates the modeling and learning processes for the two patterns. 

\subsection{Dual Attention Module}

We propose a `Dual Attention Module' to model the inter-series dependencies and intra-series variations simultaneously.
Concretely, we devise `Channel-wise self-attention' for modeling the former and `Auto-regressive self-attention' for modeling the latter.

\textbf{Inter-series dependencies modeling.} To model inter-series dependencies, follow iTranformer~\citep{LiuiTransformer}, we consider $X_{Seasonal}[i,:] \in \mathbb{R}^{1 \times D}, \quad i = 1, 2, \ldots, N$ as a token as shown in Figure \ref{fig:Two-embedding}. Subsequently, all tokens are sent into a vanilla transformer encoder for learning purposes to get $X_{Inter} \in \mathbb{R}^{N \times D}$:
\begin{align}
    Q = X_{Seasonal}W  + b, \notag  \quad 
    K = X_{Seasonal}W  + b, \\
    V = X_{Seasonal}W  + b,  \quad
    W \in \mathbb{R}^{D \times D},b \in \mathbb{R}^{1 \times D}, \notag \\
    Attn=\text{softmax }\Big( \frac{QK^{T}}{\sqrt{d_{k}}}\Big)V, \\
    H=\text{LayerNorm}(X_{Seasonal}+Attn), \notag\\
    X_{Inter}=\text{LayerNorm}(FFN(H)+H).\notag
\end{align}
We denote this process as `Channel-wise self-attention' and the generated embeddings as `Whole Series Embedding', which, while, maintaining most information of the sequence in contrast to patches or segments, is better suitable for modeling inter-series dependencies, as all semantic information of the variates are saved.

\textbf{Intra-series variations modeling.} To capture intra-series variations, we propose an advanced methodology. Herein, the initial sequence undergoes auto-regressive processing, generating tokens that, while meticulously retaining the entirety of the original information, partially emulate the dynamic variations present in raw time series.

For intra-series variations, as shown in Figure \ref{fig:Two-embedding},  we first split $X_{Seasonal}$ into $N$ individual time series $x_{s}^{i} \in \mathbb{R}^{1 \times D},\quad i = 1, 2, \ldots, N$. For each $x_{s}^{i}$, given a length $L$, we generate $S^{i} \in \mathbb{R}^{\frac{D}{L} \times D}$ tokens by cutting the given length of the sequence from the beginning and concatenate it to the end of the time series:
\begin{align}
S^{i}[j, :] = 
    x_{s}^{i}[j \cdot L : D] || x_{s}^{i}[0 : j \cdot L], \\ \notag
    \text{for } j \in \left\{0, 1, \ldots, \left\lfloor\frac{D}{L}\right\rfloor - 1\right\}.
\end{align}
Tokens generated in this process are nominated as `Auto-regressive Embedding'. This auto-regressive token generation approach enables the dynamic simulation of temporal variations within the time series. In contrast to methods involving the utilization of sampling strategies to derive sub-sequences or the partitioning of the primary time series into patches or segments, which inevitably lead to information loss, this approach maximally preserves the information of the original time series. Then we still use another vanilla transformer encoder whose weight is shared across all channels to model intra-series variations, but considering our primary interest in the temporal information of the raw sequence, we designate the raw sequence $x_{s}^{i}$ solely as Q, while utilizing the entire sequence $S^{i}$ as both K and V:
\begin{align}
    Q = x_{s}^{i}W  + b,  \notag \ \  
    K = S^{i}W  + b,  \notag \ \   
    V = S^{i}W  + b,\\
    Attn=\text{softmax }\Big( \frac{QK^{T}}{\sqrt{d_{k}}}\Big)V,\\
    H = \text{LayerNorm}(x_{s}^{i} + {Attn}), \notag\\
    X_{Intra}^{i}=\text{LayerNorm}(FFN(H)+H), \notag\\ 
    X_{Intra}=X_{Intra}^{1} || X_{Intra}^{2} || \ldots X_{Intra}^{n}. \notag
\end{align}
And the output $X_{Intra}^{i}$ of all channels is concatenated as $X_{Intra} \in \mathbb{R}^{N \times D}$. 
The ultimate output results of the seasonal part have been obtained by: 
\begin{align}
    W &\in \mathbb{R}^{D \times F},b \in \mathbb{R}^{1 \times F},\notag\\
    X_{S_{out}} &= (X_{Inter}+X_{Intra})W+b.
\end{align}

\textbf{Prediction generating.} We obtain the ultimate predictive outcomes through $\hat{Y} = X_{S_{out}}+X_{T_{out}}$.

\section{Experiments}

\subsection{Experimental Settings}
In this section, we first introduce the whole experiment settings under a fair comparison.
Secondly, we illustrate the experiment results by comparing Leddam with the eight current state-of-the-art (SOTA) methods.
Further, we conducted an ablation study to comprehensively investigate the effectiveness of the learnable convolutional decomposition module and the effectiveness of the Dual attention module.

\begin{table}[!h]
   \caption{The Statistics of the eight datasets used in our experiments.}
   \label{t:datasets}
   \begin{center}
   \large
   \resizebox{0.5\textwidth}{!}{
      \begin{tabular}{c|cccccc}
        \toprule
         {Datasets}    & {ETTh1\&2} & {ETTm1\&2} & {Traffic} & {Electricity} & {Solar-Energy} & {Weather}
         \\ \hline
         \T
         Variates    & 7        & 7        & 862     & 321         & 137             & 21      \\
         Timesteps   & 17,420   & 69,680   & 17,544  & 26,304      & 52,560         & 52,696  \\
         Granularity & 1 hour   & 5 min    & 1 hour  & 1 hour      & 10 min         & 10 min \\
         \bottomrule
      \end{tabular}
      }
   \end{center}
      \vspace{-5mm}
\end{table}
\textbf{Datasets.} We conduct extensive experiments on selected eight widely-used real-world multivariate time series forecasting datasets, including Electricity Transformer Temperature (ETTh1, ETTh2, ETTm1, and ETTm2), Electricity, Traffic, Weather used by Autoformer~\citep{Wu2021autoformer}, and Solar-Energy datasets proposed in LSTNet~\citep{Lai2018lstnet}.
For a fair comparison, we follow the same standard protocol~\citep{LiuiTransformer} and split all forecasting datasets into training, validation, and test sets by the ratio of 6:2:2 for the ETT dataset and 7:1:2 for the other datasets. The characteristics of these datasets are shown in Table \ref{t:datasets} (More can be found in the Appendix~\ref{app:datasets details}).

\begin{table*}[t]
    \centering
    \caption{Multivariate forecasting results with prediction lengths $F \in \{96, 192, 336, 720\}$ and fixed look-back length $T = 96$. Results are averaged from all prediction lengths. Full results are listed in the Appendix.}
    \label{t:avg-baseline}
    \resizebox{\textwidth}{!}{
        \begin{tabular}{c|cc|cc|cc|cc|cc|cc|cc|cc|cc}
            \toprule
            Models & \multicolumn{2}{c}{\textbf{Leddam}} & \multicolumn{2}{c}{iTransformer} & \multicolumn{2}{c}{TimesNet} & \multicolumn{2}{c}{MICN} & \multicolumn{2}{c}{DLinear} & \multicolumn{2}{c}{PatchTST} & \multicolumn{2}{c}{Crossformer} & \multicolumn{2}{c}{TiDE} & \multicolumn{2}{c}{SCINet} \\
              & \multicolumn{2}{c}{\textbf{Ours}} & \multicolumn{2}{c}{\citeyearpar{LiuiTransformer}} & \multicolumn{2}{c}{\citeyearpar{wu2022timesnet}} & \multicolumn{2}{c}{\citeyearpar{wang2023micn}} & \multicolumn{2}{c}{\citeyearpar{zeng2023dlinear}} & \multicolumn{2}{c}{\citeyearpar{Nie2022patchtst}} & \multicolumn{2}{c}{\citeyearpar{zhang2023crossformer}} & \multicolumn{2}{c}{\citeyearpar{Das2023TiDE}} & \multicolumn{2}{c}{\citeyearpar{Liu2021scinet}} \\
            \cmidrule(lr){2-3} \cmidrule(lr){4-5} \cmidrule(lr){6-7} \cmidrule(lr){8-9} \cmidrule(lr){10-11} \cmidrule(lr){12-13} \cmidrule(lr){14-15}\cmidrule(lr){16-17}\cmidrule(lr){18-19}\\
            Metric & MSE & MAE & MSE & MAE & MSE & MAE & MSE & MAE & MSE & MAE & MSE & MAE & MSE & MAE & MSE & MAE & MSE & MAE \\
            \midrule
            ETTh1 & \textcolor{red}{\textbf{0.431}} & \textcolor{red}{\textbf{0.429}} & 0.454 & 0.447 &  0.458 & 0.450 & 0.561 & 0.535 & 0.456 & 0.452 & 0.449 & 0.442 & 0.624 & 0.575 & \textcolor{blue}{\underline{0.434}} & \textcolor{blue}{\underline{0.437}} & 0.486 & 0.467 \\
            \cmidrule(lr){1-19}
            ETTh2 & \textcolor{red}{\textbf{0.373}} & \textcolor{red}{\textbf{0.399}} & \textcolor{blue}{\underline{0.383}} & \textcolor{blue}{\underline{0.407}} & 0.414 & 0.427 & 0.587 & 0.525 & 0.559 & 0.515 & 0.387 & 0.407 & 0.942 & 0.684 & 0.611 & 0.550 & 0.954 & 0.723 \\
            \cmidrule(lr){1-19}
            ETTm1 & \textcolor{red}{\textbf{0.386}} & \textcolor{red}{\textbf{0.397}} & 0.407 & 0.410 & 0.400 & 0.406 & 0.392 & 0.414 & 0.403 & 0.407 & \textcolor{blue}{\underline{0.387}} & \textcolor{blue}{\underline{0.400}} & 0.513 & 0.509 & 0.403 & 0.427 & 0.411 & 0.418 \\
            \cmidrule(lr){1-19}
            ETTm2 & \textcolor{red}{\textbf{0.281}} & \textcolor{red}{\textbf{0.325}} & 0.288 & 0.332 & 0.291 & 0.333 & 0.328 & 0.382 & 0.350 & 0.401 & \textcolor{blue}{\underline{0.283}} & \textcolor{blue}{\underline{0.327}} & 1.219 & 0.827 & 0.293 & 0.336 & 0.310 & 0.347 \\
            \cmidrule(lr){1-19}
            Electricity & \textcolor{red}{\textbf{0.169}} &  \textcolor{red}{\textbf{0.263}} & \textcolor{blue}{\underline{0.178}} & \textcolor{blue}{\underline{0.270}} & 0.192 & 0.295 & 0.187 & 0.295 & 0.212 & 0.300 & 0.216 & 0.304 & 0.244 & 0.334 & 0.251 & 0.344 & 0.268 & 0.365 \\
            \cmidrule(lr){1-19}
            Solar-Energy & \textcolor{red}{\textbf{0.230}} & \textcolor{blue}{\underline{0.264}} & \textcolor{blue}{\underline{0.233}} & \textcolor{red}{\textbf{0.262}} & 0.301 & 0.319 & 0.296 & 0.371 & 0.330 & 0.401 & 0.270 & 0.307 & 0.641 & 0.639 & 0.347 & 0.417 & 0.282 & 0.375 \\
            \cmidrule(lr){1-19}
            Traffic &  \textcolor{blue}{\underline{0.467}} &  \textcolor{blue}{\underline{0.294}} & \textcolor{red}{\textbf{0.428}} & \textcolor{red}{\textbf{0.282}} & 0.620 & 0.336 & 0.542 & 0.315 & 0.625 & 0.383 & 0.555 & 0.362 & 0.550 & 0.304 & 0.760 & 0.473 & 0.804 & 0.509 \\
            \cmidrule(lr){1-19}
            Weather & \textcolor{red}{\textbf{0.242}} & \textcolor{red}{\textbf{0.272}} & 0.258 &\textcolor{blue}{\underline{0.279}} & 0.259 & 0.287 & \textcolor{blue}{\underline{0.243}} & 0.299 & 0.265 & 0.317 & 0.259 & 0.281 & 0.259 & 0.315 & 0.271 & 0.320 & 0.292 & 0.363 \\
             \cmidrule(lr){1-19}
            $1^{st}$ Count & 7 & 6 & 1 & 2 & 0 & 0 & 0 & 0 & 0 & 0 & 0 & 0 & 0 & 0 & 0 & 0 & 0 & 0\\
            \bottomrule
        \end{tabular}
    }
\end{table*}

\textbf{Evaluation protocol.} Following TimesNet~\citep{wu2022timesnet}, we use Mean Squared Error (MSE) and Mean Absolute Error (MAE) as the core metrics for the evaluation. 
To fairly compare the forecasting performance, we follow the same evaluation protocol, where the length of the historical horizon is set as $T=96$ for all models and the prediction lengths $F \in \{96, 192, 336, 720\}$. Detailed hyperparameters of Leddam can be found in the Appendix~\ref{app:implementation}.

\textbf{Baseline setting.} We carefully choose very recently EIGHT well-acknowledged forecasting models as our baselines, including 1) Transformer-based methods: iTransformer~\citep{LiuiTransformer}, Crossformer~\citep{zhang2023crossformer}, PatchTST~\citep{Nie2022patchtst}; 2) Linear-based methods: DLinear~\citep{zeng2023dlinear}, TiDE~\citep{Das2023TiDE}; and 3) TCN-based methods: SCINet~\citep{Liu2021scinet}, MICN~\citep{wang2023micn}, TimesNet~\citep{wu2022timesnet}.

\subsection{Experiments Results}

\textbf{Quantitative comparison.} Comprehensive forecasting results are listed in Table~\ref{t:avg-baseline} with the best bold in red and the second underlined in blue.  We leave full forecasting results in APPENDIX~\ref{app:full results} to save place. The lower MSE/MAE indicates the better prediction result. It is unequivocally evident that Leddam has demonstrated superior predictive performance across all datasets except Traffic, in which iTransformer gets the best forecasting performance. Like PatchTST and iTransformer, Leddam employs a vanilla transformer encoder as its backbone, devoid of any structural modifications. It is noteworthy, however, that these three models consistently exhibit superior performance across all datasets. This at least partially indicates that, compared to intricately designed model architectures, superior representation of raw time series features, such as `Whole Series Embedding' used in iTransformer and Leddam along with PatchTST's patches, may indeed constitute the pivotal factors for achieving more efficient time-series predictions. It is noteworthy that among these three models, PacthTST employs a channel-independent design, exclusively addressing intra-series variations without considering inter-series dependencies. iTransformer utilizes channel-wise self-attention to model inter-series dependencies but falls short of adequately capturing intra-series variations. Compared to both, the proposed Leddam incorporates `Whole Series Embedding' and `Auto-regressive Embedding' to model inter-series dependencies and intra-series variations. Consequently, Leddam demonstrates superior performance across various datasets. However, these three models still maintain a leading position over other models across most datasets. This aligns with our hypothesis that appropriately modeling inter-series dependencies and intra-series variations in multivariate time series is key for achieving more precise forecasting.

\begin{table*}[t]
\centering
\caption{Ablation of model structure across five datasets with prediction lengths $F = 96$, and input length $T = 96$.}
\label{t:ablation of structure}
\resizebox{0.7\textwidth}{!}{
\begin{tabular}{ccccccccccc}
\toprule
\multirow{2}{*}{Models} & \multicolumn{2}{c}{ETTm1} & \multicolumn{2}{c}{Electricity} & \multicolumn{2}{c}{Traffic} & \multicolumn{2}{c}{Weather} & \multicolumn{2}{c}{Solar-Energy} \\
\cmidrule(lr){2-3} \cmidrule(lr){4-5} \cmidrule(lr){6-7} \cmidrule(lr){8-9} \cmidrule(lr){10-11} 
& MSE & MAE & MSE & MAE & MSE & MAE & MSE & MAE & MSE & MAE \\
\midrule
w/o All & 0.350 & 0.368 & 0.197 & 0.275 & 0.644 & 0.389 & 0.195 & 0.235 & 0.306 & 0.330 \\
\cmidrule(lr){1-11}
w/o Auto & 0.337 & 0.371  & 0.148 & 0.242 & 0.438 & 0.288 & 0.168 & 0.212 & 0.211 & 0.253 \\
\cmidrule(lr){1-11}
w/o Channel  & 0.335 & 0.369  & 0.152 & 0.242 & 0.467 & 0.285 & 0.167 & 0.212 & 0.226 & 0.263 \\
\cmidrule(lr){1-11}
\textbf{Leddam (Ours)} & \textbf{0.320} & \textbf{0.359} & \textbf{0.139} & \textbf{0.233} & \textbf{0.424} & \textbf{0.269} & \textbf{0.158} & \textbf{0.203} & \textbf{0.202} & \textbf{0.240} \\
\bottomrule
\end{tabular}
}
\end{table*}

\begin{table*}[t]
\centering
\caption{Predictive performance comparison of moving average kernel and trainable 1D convolutional kernel across four datasets. The prediction horizon is uniformly set at $F = 720$, while the input length $T = 96$.}
\label{t:ablation_of_kernel}
\resizebox{0.66\textwidth}{!}{
\begin{tabular}{ccccccccccc}
\toprule
\multirow{2}{*}{Design} & \multicolumn{2}{c}{ETTh2} & \multicolumn{2}{c}{ETTm2} & \multicolumn{2}{c}{Electricity} & \multicolumn{2}{c}{Traffic} & \multicolumn{2}{c}{Solar-Energy} \\
\cmidrule(lr){2-3} \cmidrule(lr){4-5} \cmidrule(lr){6-7} \cmidrule(lr){8-9} \cmidrule(lr){10-11} 
& MSE & MAE & MSE & MAE & MSE & MAE & MSE & MAE & MSE & MAE \\
\midrule
DLinear & 0.831  & 0.657 & 0.554 & 0.522 & 0.245 & 0.333 & 0.645 & 0.394 & 0.356 & 0.413 \\
\cmidrule(lr){1-11}
LD\_UTL  & 0.742 & 0.607 & 0.541 & 0.505  & 0.220 & 0.311 & 0.602 & 0.378 & 0.333 & 0.397 \\
\cmidrule(lr){1-11}
\textbf{LD\_TL} & \textbf{0.684} & \textbf{0.576} & \textbf{0.521} & \textbf{0.488} & \textbf{0.209} & \textbf{0.302} & \textbf{0.584} & \textbf{0.345} & \textbf{0.313} & \textbf{0.376} \\
\bottomrule
\end{tabular}
}
\end{table*}

\subsection{Model Analysis}
\textbf{Ablation study of dual attention module}  We conducted ablation experiments across five datasets, including ETTh1, Traffic, Electricity, Solar-Energy, and Weather, to validate the performance enhancement introduced by our `Auto-regressive self-attention' and `Channel-wise self-attention' components. `Channel' means we only used `Channel-wise self-attention'. `Auto' means we only used `Auto-regressive self-attention'. `w/o All' means we simply replace the `Auto-regressive self-attention' and `Channel-wise self-attention' components with a linear layer.  

We observe substantial performance improvements in Table~\ref{t:ablation of structure} introduced by the `Auto-regressive self-attention' and `Channel-wise self-attention' components. `Auto-regressive self-attention'  brings an average of 19.02\% of MSE to decrease across five datasets compared to using linear layer, and `Channel-wise self-attention' achieves 21.09\% improvement.
Moreover, their synergistic integration yields further enhancements in model performance, getting an average 25.03\% of MSE decrease, reaching an optimal level. 
This attests to the efficacy of both design elements and once again validates our hypothesis, namely, appropriately modeling inter-series dependencies and intra-series variations in multivariate time series can yield better predictive performance.

\textbf{Superiority of Learnable Decomposition Module over Moving Average Kernel.} To better emphasize the advantages of our Learnable Decomposition Module with weights initialized using a Gaussian distribution, in comparison to conventional Moving Average Kernel, we conducted an extensive experiment. Given that DLinear arguably represents the most prominent instantiation utilizing a Moving Average Kernel for trend information extraction, and achieves performance comparable to other state-of-the-art methods with the mere use of two simple linear layers, we select it as our baseline. For comparison, we substitute its Moving Average Kernel with our Learnable Decomposition Module, denoting the modified model as LD\_TL if the kernel is set to trainable, else LD\_UTL if the kernel is set to untrainable. 

In comparison to a simple Moving Average Kernel, the Learnable Decomposition Module, as depicted in Table~\ref{t:ablation_of_kernel}, consistently exhibits superior predictive performance across all four datasets regardless of its trainability. The untrainable version of the 1D convolutional kernel brings an average  7.28\% of MSE decrease across five datasets compared to using a Moving Average Kernel, while the trainable version gives 11.98\%. The obtained results conclusively demonstrate the superior efficacy of our Learnable Decomposition Module over a simple Moving Average Kernel, and the trainability of the kernel plays a significant role in its adaptability. We leave further analysis to APPENDIX~\ref{app:decom results visualization}-\ref{app:ms_deom}.

\textbf{Decomposition result analysis.} 
To further investigate why our Learnable Decomposition Module (LD) is a better time series decomposition solution than Moving Average Kernel (MOV), we conducted the following analysis on the seasonal part and trend part obtained. 

Since the seasonal part represents repetitive patterns in the raw sequence, a good seasonal part should capture all major frequencies in the raw sequence. We separately calculated the amplitude similarity of the dominant frequencies (top 25\%) between the seasonal part obtained by each method and the raw sequence, denoted as \textbf{FFT}. A better decomposition strategy should yield a seasonal part with higher similarity to the dominant frequencies of the raw sequence.

A good trend part should effectively capture the trend changes in the original sequence. Thus, we employed Dynamic Time Warping (DTW) \citep{Müller2007dtw} to compute the similarity between the raw sequence and the trend parts obtained from the two decompositions, denoted as \textbf{DTW}. A superior decomposition strategy should result in a trend part with higher DTW similarity to the raw sequence.

We selected \textbf{the final variate of each dataset}. The variates used are as follows.\vspace{-0.15cm}
\begin{itemize}
 \item \textbf{ETT Dataset}: Oil Temperature (OT) every hour or 15 minutes.\vspace{-0.15cm}
 \item \textbf{Electricity}: hourly electricity consumption of the 321st (last) user.\vspace{-0.15cm}
 \item  \textbf{Solar-Energy}: solar power production every 10 minutes of the 137th (last) PV plant.\vspace{-0.15cm}
 \item  \textbf{Traffic}: hourly road occupancy rates measured by the 862nd (last) sensor.\vspace{-0.15cm}
 \item  \textbf{Weather}: CO2 (ppm) collected every 10 minutes.\vspace{-0.15cm}
\end{itemize}
  
LD is pre-trained on task: input-96-forecast-720. To avoid cherry-picking, both metrics are computed by averaging the results calculated over the entire test dataset.

\begin{table}[t]
\centering
\caption{Decomposition result analysis. Comparison of DTW and FFT for LD and MOV across eight datasets.}
\label{t:kernel_comparison}
\resizebox{0.4\textwidth}{!}{
\begin{tabular}{c|c|c|c|c}
\toprule
Kernel & \multicolumn{2}{c}{\textbf{LD (Ours)}} & \multicolumn{2}{c}{MOV} \\
\midrule
Dataset/Metric & DTW & FFT & DTW & FFT \\
\midrule
Electricity & \textbf{0.643} & \textbf{0.942} & 0.618 & 0.931 \\ \midrule
Solar\_Energy & \textbf{0.910} & \textbf{0.781} & 0.873 & 0.734 \\  \midrule
Traffic & \textbf{0.603} & \textbf{0.993} & 0.563 & 0.991 \\ \midrule
Weather & \textbf{0.858} & \textbf{0.760} & 0.846 & 0.691 \\ \midrule
ETTh1 & \textbf{0.741} & \textbf{0.892} & 0.724 & 0.852 \\ \midrule
ETTh2 & \textbf{0.675} & \textbf{0.900} & 0.652 & 0.887 \\ \midrule
ETTm1 & \textbf{0.821} & \textbf{0.754} & 0.808 & 0.717 \\ \midrule
ETTm2 & \textbf{0.925} & \textbf{0.894} & 0.908 & 0.759 \\
\bottomrule
\end{tabular}
}
\end{table}

The DTW and FFT of LD are consistently superior to that of the MOV across all eight datasets in Table~\ref{t:kernel_comparison}. This demonstrates that LD is a better time series decomposition method compared to MOV.

\begin{figure}[t]
  \centering
  \includegraphics[width=1.0\linewidth]{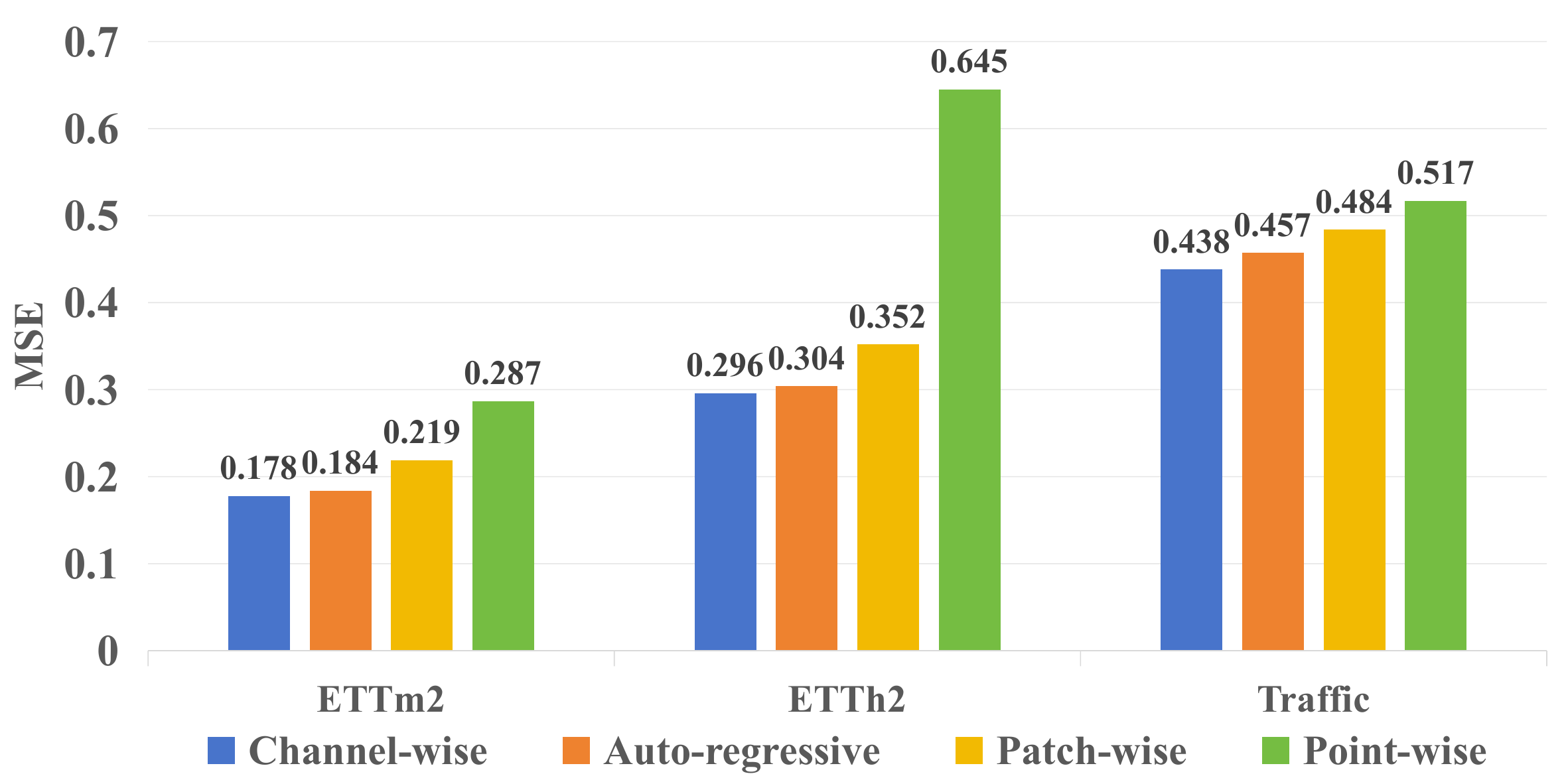}
  \caption{Predictive performance comparison(MSE) of `Channel-wise self-attention', `Auto-regressive self-attention', `Patch-wise self-attention', and `Point-wise self-attention' across ETTh2, ETTm2 and Traffic datasets. The prediction horizon is uniformly set at $F = 96$, while the input length $T = 96$.}
  \label{fig: modeling analysis}
\end{figure}

\textbf{Analysis of different attention mechanisms in inter-series dependencies modeling.} We devised a comprehensive experiment to investigate three prevalent approaches for modeling inter-series dependencies: `Channel-wise self-attention', `Point-wise self-attention', and `Patch-wise self-attention'. The first considers the entire sequence of a variate as a token~\citep{LiuiTransformer}, the second regards distinct variables at the same timestamp as tokens~\citep{Zhou2020informer,Zhou2022fedformer}, and the third treats patches of the raw sequence as tokens~\citep{zhang2023crossformer,Nie2022patchtst}.
Specifically, we eliminate the `Auto-regressive self-attention' branch from the original Leddam structure to concentrate on inter-series dependencies modeling. 
We sequentially employ `Channel-wise self-attention', `Point-wise self-attention', and `Patch-wise self-attention'. 
In Figure~\ref{fig: modeling analysis}, it is readily apparent that compared to `Point-wise self-attention', and `Patch-wise self-attention', `Channel-wise self-attention' exhibits superior predictive performance, implies a much better inter-series dependencies modeling ability. 

\textbf{Analysis of different attention mechanisms in intra-series variations modeling.}
Similarly, we replace the `Channel-wise self-attention' branch from the original Leddam structure and use `Point-wise self-attention', `Patch-wise self-attention', and `Auto-regressive self-attention' for intra-series variations modeling.
The superiority of the `Auto-regressive self-attention' architecture is proved by the experimental results in Figure~\ref{fig: modeling analysis}.

\subsection{Learnable Decomposition Generalization Analysis}
In this subsection, we investigate the generalizability of the learnable decomposition module of \textbf{Leddam} by plugging it into other different kinds of models.

\begin{table*}[!h]
    \centering
    \caption{Improvements of LD over different models with prediction lengths $F = 96$, and fixed lookback length $T = 96$. LD means Learnable decomposition.}
    \label{t:boosting}
    \resizebox{0.80\textwidth}{!}{
    \begin{tabular}{c|c|cc|cc|cc|cc|cc}
        \toprule
        & Models & \multicolumn{2}{c}{LightTS} & \multicolumn{2}{c}{LSTM} & \multicolumn{2}{c}{SCINet} & \multicolumn{2}{c}{Informer} & \multicolumn{2}{c}{Transformer} \\
        \cmidrule(lr){3-4} \cmidrule(lr){5-6} \cmidrule(lr){7-8} \cmidrule(lr){9-10} \cmidrule(lr){11-12}
        & Metric & MSE & MAE & MSE & MAE & MSE & MAE & MSE & MAE & MSE & MAE\\
        \midrule
         \multirow{3}{*}{ETTh2} & Original & 0.439 & 0.464 & 1.537 & 0.987 & 0.784 & 0.680 & 2.234 & 1.239 & 1.528 & 1.029 \\
                        & + \textbf{LD} & \textbf{0.370} & \textbf{0.413} & \textbf{0.354} & \textbf{0.395} & \textbf{0.446} & \textbf{0.469} & \textbf{1.086} & \textbf{0.804} & \textbf{0.645} & \textbf{0.635} \\
                        \cmidrule(lr){2-12}
                        & \textbf{Improvement} & \textcolor{red}{\textbf{15.78\%}} & \textcolor{red}{\textbf{10.97\%}} & \textcolor{red}{\textbf{76.97\%}} & \textcolor{red}{\textbf{59.99\%}}  & \textcolor{red}{\textbf{43.12\%}} & \textcolor{red}{\textbf{31.04\%}} & \textcolor{red}{\textbf{51.40\%}} & \textcolor{red}{\textbf{35.10\%}} & \textcolor{red}{\textbf{57.77\%}} & \textcolor{red}{\textbf{38.30\%}} \\
        
        \cmidrule(lr){2-12}
        \addlinespace
         \multirow{3}{*}{ETTm2} & Original & 0.249 & 0.345 & 1.009 & 0.797  & 0.302 & 0.406 & 0.368 & 0.475 & 0.295 & 0.418 \\
                        & + \textbf{LD} & \textbf{0.201} & \textbf{0.293} & \textbf{0.199} & \textbf{0.299}  & \textbf{0.220} & \textbf{0.322} & \textbf{0.302} & \textbf{0.402} & \textbf{0.287} & \textbf{0.399} \\
                        \cmidrule(lr){2-12}
                        & \textbf{Improvement} & \textcolor{red}{\textbf{19.05\%}} & \textcolor{red}{\textbf{15.00\%}} & \textcolor{red}{\textbf{80.28\%}} & \textcolor{red}{\textbf{62.48\%}}  & \textcolor{red}{\textbf{27.03\%}} & \textcolor{red}{\textbf{20.71\%}} & \textcolor{red}{\textbf{17.91\%}} & \textcolor{red}{\textbf{15.37\%}} & \textcolor{red}{\textbf{2.58\%}} & \textcolor{red}{\textbf{4.45\%}} \\
        \cmidrule(lr){2-12}
        \addlinespace
         \multirow{3}{*}{Weather} & Original & 0.166 & 0.234 & 0.221 & 0.302 & 0.214 & 0.290 & 0.201 & 0.285 & 0.167 & 0.248 \\
                          & + \textbf{LD} & \textbf{0.165} & \textbf{0.232} & \textbf{0.186} & \textbf{0.258}  & \textbf{0.181} & \textbf{0.261} & \textbf{0.177} & \textbf{0.259} & \textbf{0.164} & \textbf{0.242} \\
                          \cmidrule(lr){2-12}
                          & \textbf{Improvement} & \textcolor{red}{\textbf{0.60\%}} & \textcolor{red}{\textbf{0.77\%}} & \textcolor{red}{\textbf{15.84\%}} & \textcolor{red}{\textbf{14.57\%}}  & \textcolor{red}{\textbf{15.33\%}} & \textcolor{red}{\textbf{9.96\%}} & \textcolor{red}{\textbf{11.72\%}} & \textcolor{red}{\textbf{9.00\%}} & \textcolor{red}{\textbf{1.32\%}} & \textcolor{red}{\textbf{2.14\%}} \\

        \cmidrule(lr){2-12}
        \addlinespace
        \multirow{3}{*}{Electricity} &  Original & 0.233 & 0.337 & 0.305 & 0.384  & 0.247 & 0.345  & 0.374 & 0.441 & 0.275 & 0.368 \\
                             & + \textbf{LD} & \textbf{0.194} & \textbf{0.297} & \textbf{0.159} & \textbf{0.263}  & \textbf{0.201} & \textbf{0.306} & \textbf{0.220} & \textbf{0.321} & \textbf{0.166} & \textbf{0.273} \\
                             \cmidrule(lr){2-12}
                             & \textbf{Improvement} & \textcolor{red}{\textbf{16.74\%}} & \textcolor{red}{\textbf{11.79\%}} & \textcolor{red}{\textbf{47.80\%}} & \textcolor{red}{\textbf{31.44\%}} & \textcolor{red}{\textbf{18.62\%}} & \textcolor{red}{\textbf{11.30\%}}& \textcolor{red}{\textbf{41.16\%}} & \textcolor{red}{\textbf{27.18\%}} & \textcolor{red}{\textbf{39.64\%}} & \textcolor{red}{\textbf{25.81\%}} \\

        \cmidrule(lr){2-12}
        \addlinespace
          \multirow{3}{*}{Traffic} & Original & 0.640 & 0.405 & 0.680 & 0.375  & 0.668 & 0.427 & 0.806 & 0.453 & 0.739 & 0.397 \\
                 &+ \textbf{LD} & \textbf{0.594} & \textbf{0.378} & \textbf{0.531} & \textbf{0.338}  & \textbf{0.590} & \textbf{0.373} & \textbf{0.640} & \textbf{0.415} & \textbf{0.517} & \textbf{0.312} \\
                 \cmidrule(lr){2-12}
                 & \textbf{Improvement} & \textcolor{red}{\textbf{7.16\%}} & \textcolor{red}{\textbf{6.60\%}} & \textcolor{red}{\textbf{21.91\%}} & \textcolor{red}{\textbf{9.87\%}}  & \textcolor{red}{\textbf{11.64\%}} & \textcolor{red}{\textbf{12.65\%}} & \textcolor{red}{\textbf{20.56\%}} & \textcolor{red}{\textbf{8.45\%}} & \textcolor{red}{\textbf{30.04\%}} & \textcolor{red}{\textbf{21.41\%}} \\

        \bottomrule
    \end{tabular}%
    }
\end{table*}

\textbf{Model selection and experimental setting.} To achieve this objective, we conduct experiments across a spectrum of representative time series forecasting model structures, including (1) Transformer-based methods: Informer~\citep{Zhou2020informer}, Transformer~\citep{Vaswani2017transformer}; (2) Linear-based methods: LightTS~\citep{Zhang2022lightts}; and (3) TCN-based methods: SCINet~\cite{Liu2021scinet}; (4) RNN-based methods: LSTM~\citep{Hochreiter1997lstm}. We standardize the input length $T$ to 96, and similarly, the prediction length $F$ is uniformly set to 96. Subsequently, comparative experiments were conducted on five datasets: ETTh2, ETTm2, Weather, Electricity, and Traffic. Specifically, we sequentially replaced the `Auto-regressive self-attention' and `Channel-wise self-attention' components in \textbf{Leddam} with each model. The comparative analysis was performed to assess the predictive performance before and after the incorporation of \textbf{Leddam}, comparing the original models with the augmented counterparts. And more results can be found in APPENDIX~\ref{app:generalization}.

\textbf{Quantitative results.} In Table~\ref{t:boosting}, it is apparent that the incorporation of the \textbf{Leddam} structure leads to a notably substantial enhancement in the predictive performance of various models, even with the introduction of only a single linear layer.
Specifically, LightTS demonstrates an average MSE reduction of 11.87\% across five datasets, other models are LSTM: 48.56\%, SCINet: 23.15\%, Informer: 31.72\%, and Transformer: 26.27\%. Particularly noteworthy is the performance enhancement observed in the classical LSTM model, where the MSE experiences a remarkable decrease of 76.97\% and 80.28\% on ETTh2 and ETTm2, respectively, a profoundly surprising result. This unequivocally substantiates the generality of the \textbf{Leddam} structure.

\section{Conclusion}
Given the non-linear and intricate trend characteristics inherent in real-world time series data, this paper formulates a learnable convolution kernel as an improvement over the simple moving average kernel for time series decomposition. 
The Gaussian initialization and adaptable properties enable it to better align with the nuances of real-time series data, resulting in a more contextually fitting decomposition. 
Additionally, we present the dual attention module, incorporating both channel-wise self-attention and autoregressive self-attention. This innovative design facilitates the simultaneous capture of inter-series dependencies and intra-series variations with precision. Experimentally, our method achieves state-of-the-art performance and demonstrates remarkable framework generality, as supported by compelling analyses. 
In the future, we aim to optimize the application of our learnable convolutional kernel in the context of series decomposition.

\section*{Acknowledgements}
This work was partially supported by the Start-up Fund of The Hong Kong Polytechnic University (No. P0045999) and the Seed Fund of the Research Institute for Smart Ageing (No. P0050946). 
We thank anonymous reviewers for their comments and suggestions. 

\section*{Impact Statement}
This paper presents work whose goal is to advance the field of  Machine Learning. There are many potential societal consequences of our work, none of which we feel must be specifically highlighted here.

\bibliography{DEFT}

\begin{thebibliography}{38}
\providecommand{\natexlab}[1]{#1}
\providecommand{\url}[1]{\texttt{#1}}
\expandafter\ifx\csname urlstyle\endcsname\relax
  \providecommand{\doi}[1]{doi: #1}\else
  \providecommand{\doi}{doi: \begingroup \urlstyle{rm}\Url}\fi

\bibitem[Mü(2007)]{Müller2007dtw}
\emph{Dynamic Time Warping}, pp.\  69--84.
\newblock Springer Berlin Heidelberg, Berlin, Heidelberg, 2007.
\newblock ISBN 978-3-540-74048-3.
\newblock \doi{10.1007/978-3-540-74048-3_4}.
\newblock URL \url{https://doi.org/10.1007/978-3-540-74048-3_4}.

\bibitem[Anonymous(2024{\natexlab{a}})]{anonymous2024fits}
Anonymous.
\newblock {FITS}: Modeling time series with \$10k\$ parameters.
\newblock In \emph{The Twelfth International Conference on Learning Representations}, 2024{\natexlab{a}}.
\newblock URL \url{https://openreview.net/forum?id=bWcnvZ3qMb}.

\bibitem[Anonymous(2024{\natexlab{b}})]{anonymous2024moderntcn}
Anonymous.
\newblock Modern{TCN}: A modern pure convolution structure for general time series analysis.
\newblock In \emph{The Twelfth International Conference on Learning Representations}, 2024{\natexlab{b}}.
\newblock URL \url{https://openreview.net/forum?id=vpJMJerXHU}.

\bibitem[Anonymous(2024{\natexlab{c}})]{anonymous2024timemixer}
Anonymous.
\newblock Timemixer: Decomposable multiscale mixing for time series forecasting.
\newblock In \emph{The Twelfth International Conference on Learning Representations}, 2024{\natexlab{c}}.
\newblock URL \url{https://openreview.net/forum?id=7oLshfEIC2}.

\bibitem[Das et~al.(2023)Das, Kong, Leach, Mathur, Sen, and Yu]{Das2023TiDE}
Das, A., Kong, W., Leach, A., Mathur, S.~K., Sen, R., and Yu, R.
\newblock Long-term forecasting with {TiDE}: Time-series dense encoder.
\newblock \emph{Transactions on Machine Learning Research}, 2023.
\newblock ISSN 2835-8856.
\newblock URL \url{https://openreview.net/forum?id=pCbC3aQB5W}.

\bibitem[Dong et~al.(2023)Dong, Wu, Zhang, Zhang, Wang, and Long]{dong2023simmtm}
Dong, J., Wu, H., Zhang, H., Zhang, L., Wang, J., and Long, M.
\newblock Sim{MTM}: A simple pre-training framework for masked time-series modeling.
\newblock In \emph{Thirty-seventh Conference on Neural Information Processing Systems}, 2023.
\newblock URL \url{https://openreview.net/forum?id=ginTcBUnL8}.

\bibitem[Dosovitskiy et~al.(2021)Dosovitskiy, Beyer, Kolesnikov, Weissenborn, Zhai, Unterthiner, Dehghani, Minderer, Heigold, Gelly, Uszkoreit, and Houlsby]{Dosovitskiy2020vit}
Dosovitskiy, A., Beyer, L., Kolesnikov, A., Weissenborn, D., Zhai, X., Unterthiner, T., Dehghani, M., Minderer, M., Heigold, G., Gelly, S., Uszkoreit, J., and Houlsby, N.
\newblock An image is worth 16x16 words: Transformers for image recognition at scale.
\newblock In \emph{International Conference on Learning Representations}, 2021.
\newblock URL \url{https://openreview.net/forum?id=YicbFdNTTy}.

\bibitem[Han et~al.(2023)Han, Ye, and Zhan]{Han2023ci1}
Han, L., Ye, H.-J., and Zhan, D.-C.
\newblock The capacity and robustness trade-off: Revisiting the channel independent strategy for multivariate time series forecasting.
\newblock Apr 2023.

\bibitem[Hochreiter \& Schmidhuber(1997)Hochreiter and Schmidhuber]{Hochreiter1997lstm}
Hochreiter, S. and Schmidhuber, J.
\newblock Long short-term memory.
\newblock \emph{Neural Computation}, pp.\  1735–1780, Nov 1997.
\newblock \doi{10.1162/neco.1997.9.8.1735}.
\newblock URL \url{http://dx.doi.org/10.1162/neco.1997.9.8.1735}.

\bibitem[Kim et~al.(2022)Kim, Kim, Tae, Park, Choi, and Choo]{Kim_revin}
Kim, T., Kim, J., Tae, Y., Park, C., Choi, J.-H., and Choo, J.
\newblock Reversible instance normalization for accurate time-series forecasting against distribution shift.
\newblock In \emph{International Conference on Learning Representations}, 2022.
\newblock URL \url{https://openreview.net/forum?id=cGDAkQo1C0p}.

\bibitem[Kitaev et~al.(2020)Kitaev, Kaiser, and Levskaya]{Kitaev2020reformer}
Kitaev, N., Kaiser, L., and Levskaya, A.
\newblock Reformer: The efficient transformer.
\newblock In \emph{International Conference on Learning Representations}, 2020.
\newblock URL \url{https://openreview.net/forum?id=rkgNKkHtvB}.

\bibitem[Lai et~al.(2018)Lai, Chang, Yang, and Liu]{Lai2018lstnet}
Lai, G., Chang, W.-C., Yang, Y., and Liu, H.
\newblock Modeling long-and short-term temporal patterns with deep neural networks.
\newblock In \emph{The 41st international ACM SIGIR conference on research \& development in information retrieval}, pp.\  95--104, 2018.

\bibitem[Li et~al.(2019)Li, Jin, Xuan, Zhou, Chen, Wang, and Yan]{Li2019logtrans}
Li, S., Jin, X., Xuan, Y., Zhou, X., Chen, W., Wang, Y.-X., and Yan, X.
\newblock Enhancing the locality and breaking the memory bottleneck of transformer on time series forecasting.
\newblock In Wallach, H., Larochelle, H., Beygelzimer, A., d\textquotesingle Alch\'{e}-Buc, F., Fox, E., and Garnett, R. (eds.), \emph{Advances in Neural Information Processing Systems}, volume~32. Curran Associates, Inc., 2019.
\newblock URL \url{https://proceedings.neurips.cc/paper_files/paper/2019/file/6775a0635c302542da2c32aa19d86be0-Paper.pdf}.

\bibitem[Li et~al.(2023)Li, Qi, Li, and Xu]{Li2023ci2}
Li, Z., Qi, S., Li, Y., and Xu, Z.
\newblock Revisiting long-term time series forecasting: An investigation on linear mapping.
\newblock \emph{arXiv preprint arXiv:2305.10721}, 2023.

\bibitem[LIU et~al.(2022)LIU, Zeng, Chen, Xu, LAI, Ma, and Xu]{Liu2021scinet}
LIU, M., Zeng, A., Chen, M., Xu, Z., LAI, Q., Ma, L., and Xu, Q.
\newblock {SCIN}et: Time series modeling and forecasting with sample convolution and interaction.
\newblock In Oh, A.~H., Agarwal, A., Belgrave, D., and Cho, K. (eds.), \emph{Advances in Neural Information Processing Systems}, 2022.
\newblock URL \url{https://openreview.net/forum?id=AyajSjTAzmg}.

\bibitem[Liu et~al.(2022{\natexlab{a}})Liu, Yu, Liao, Li, Lin, Liu, and Dustdar]{Liu2021pyraformer}
Liu, S., Yu, H., Liao, C., Li, J., Lin, W., Liu, A.~X., and Dustdar, S.
\newblock Pyraformer: Low-complexity pyramidal attention for long-range time series modeling and forecasting.
\newblock In \emph{International Conference on Learning Representations}, 2022{\natexlab{a}}.
\newblock URL \url{https://openreview.net/forum?id=0EXmFzUn5I}.

\bibitem[Liu et~al.(2022{\natexlab{b}})Liu, Wu, Wang, and Long]{liu2022non}
Liu, Y., Wu, H., Wang, J., and Long, M.
\newblock Non-stationary transformers: Exploring the stationarity in time series forecasting.
\newblock In Oh, A.~H., Agarwal, A., Belgrave, D., and Cho, K. (eds.), \emph{Advances in Neural Information Processing Systems}, 2022{\natexlab{b}}.
\newblock URL \url{https://openreview.net/forum?id=ucNDIDRNjjv}.

\bibitem[Liu et~al.(2023)Liu, Li, Wang, and Long]{Liu2023koopa}
Liu, Y., Li, C., Wang, J., and Long, M.
\newblock Koopa: Learning non-stationary time series dynamics with koopman predictors.
\newblock In \emph{Thirty-seventh Conference on Neural Information Processing Systems}, 2023.
\newblock URL \url{https://openreview.net/forum?id=A4zzxu82a7}.

\bibitem[Liu et~al.(2024)Liu, Hu, Zhang, Wu, Wang, and Long]{LiuiTransformer}
Liu, Y., Hu, T., Zhang, H., Wu, H., Wang, S., and Long, M.
\newblock itransformer: Inverted transformers are effective for time series forecasting.
\newblock In \emph{The Twelfth International Conference on Learning Representations}, 2024.
\newblock URL \url{https://openreview.net/forum?id=JePfAI8fah}.

\bibitem[Ni et~al.(2023)Ni, Yu, Liu, Li, and Lin]{Ni2023basiformer}
Ni, Z., Yu, H., Liu, S., Li, J., and Lin, W.
\newblock Basisformer: Attention-based time series forecasting with learnable and interpretable basis.
\newblock In \emph{Thirty-seventh Conference on Neural Information Processing Systems}, 2023.
\newblock URL \url{https://openreview.net/forum?id=xx3qRKvG0T}.

\bibitem[Nie et~al.(2023)Nie, Nguyen, Sinthong, and Kalagnanam]{Nie2022patchtst}
Nie, Y., Nguyen, N.~H., Sinthong, P., and Kalagnanam, J.
\newblock A time series is worth 64 words: Long-term forecasting with transformers.
\newblock In \emph{The Eleventh International Conference on Learning Representations}, 2023.
\newblock URL \url{https://openreview.net/forum?id=Jbdc0vTOcol}.

\bibitem[Rangapuram et~al.(2018)Rangapuram, Seeger, Gasthaus, Stella, Wang, and Januschowski]{Rangapuram2018dpstate}
Rangapuram, S.~S., Seeger, M.~W., Gasthaus, J., Stella, L., Wang, Y., and Januschowski, T.
\newblock Deep state space models for time series forecasting.
\newblock In Bengio, S., Wallach, H., Larochelle, H., Grauman, K., Cesa-Bianchi, N., and Garnett, R. (eds.), \emph{Advances in Neural Information Processing Systems}, volume~31. Curran Associates, Inc., 2018.
\newblock URL \url{https://proceedings.neurips.cc/paper_files/paper/2018/file/5cf68969fb67aa6082363a6d4e6468e2-Paper.pdf}.

\bibitem[Shao et~al.(2022)Shao, Zhang, Wang, Wei, and Xu]{Shao2022trafgnn}
Shao, Z., Zhang, Z., Wang, F., Wei, W., and Xu, Y.
\newblock Spatial-temporal identity: A simple yet effective baseline for multivariate time series forecasting.
\newblock In \emph{Proceedings of the 31st ACM International Conference on Information \& Knowledge Management}, pp.\  4454--4458, 2022.

\bibitem[Taylor \& Letham(2018)Taylor and Letham]{Taylor_2018_distribution}
Taylor, S.~J. and Letham, B.
\newblock Forecasting at scale.
\newblock \emph{The American Statistician}, 72\penalty0 (1):\penalty0 37--45, 2018.
\newblock \doi{10.1080/00031305.2017.1380080}.
\newblock URL \url{https://doi.org/10.1080/00031305.2017.1380080}.

\bibitem[Vaswani et~al.(2017)Vaswani, Shazeer, Parmar, Uszkoreit, Jones, Gomez, Kaiser, and Polosukhin]{Vaswani2017transformer}
Vaswani, A., Shazeer, N., Parmar, N., Uszkoreit, J., Jones, L., Gomez, A.~N., Kaiser, L.~u., and Polosukhin, I.
\newblock Attention is all you need.
\newblock In Guyon, I., Luxburg, U.~V., Bengio, S., Wallach, H., Fergus, R., Vishwanathan, S., and Garnett, R. (eds.), \emph{Advances in Neural Information Processing Systems}, volume~30. Curran Associates, Inc., 2017.
\newblock URL \url{https://proceedings.neurips.cc/paper_files/paper/2017/file/3f5ee243547dee91fbd053c1c4a845aa-Paper.pdf}.

\bibitem[Wang et~al.(2023)Wang, Peng, Huang, Wang, Chen, and Xiao]{wang2023micn}
Wang, H., Peng, J., Huang, F., Wang, J., Chen, J., and Xiao, Y.
\newblock {MICN}: Multi-scale local and global context modeling for long-term series forecasting.
\newblock In \emph{The Eleventh International Conference on Learning Representations}, 2023.
\newblock URL \url{https://openreview.net/forum?id=zt53IDUR1U}.

\bibitem[Wu et~al.(2024)Wu, Hu, Li, Chen, and Liu]{wu2024stanhop}
Wu, D., Hu, J. Y.-C., Li, W., Chen, B.-Y., and Liu, H.
\newblock {ST}anhop: Sparse tandem hopfield model for memory-enhanced time series prediction.
\newblock In \emph{The Twelfth International Conference on Learning Representations}, 2024.
\newblock URL \url{https://openreview.net/forum?id=6iwg437CZs}.

\bibitem[Wu et~al.(2021)Wu, Xu, Wang, and Long]{Wu2021autoformer}
Wu, H., Xu, J., Wang, J., and Long, M.
\newblock Autoformer: Decomposition transformers with auto-correlation for long-term series forecasting.
\newblock In Ranzato, M., Beygelzimer, A., Dauphin, Y., Liang, P., and Vaughan, J.~W. (eds.), \emph{Advances in Neural Information Processing Systems}, volume~34, pp.\  22419--22430. Curran Associates, Inc., 2021.
\newblock URL \url{https://proceedings.neurips.cc/paper_files/paper/2021/file/bcc0d400288793e8bdcd7c19a8ac0c2b-Paper.pdf}.

\bibitem[Wu et~al.(2023)Wu, Hu, Liu, Zhou, Wang, and Long]{wu2022timesnet}
Wu, H., Hu, T., Liu, Y., Zhou, H., Wang, J., and Long, M.
\newblock {TimesNet}: Temporal 2d-variation modeling for general time series analysis.
\newblock In \emph{The Eleventh International Conference on Learning Representations}, 2023.
\newblock URL \url{https://openreview.net/forum?id=ju_Uqw384Oq}.

\bibitem[Yi et~al.(2023{\natexlab{a}})Yi, Zhang, Fan, He, Hu, Wang, An, Cao, and Niu]{yi2023fouriergnn}
Yi, K., Zhang, Q., Fan, W., He, H., Hu, L., Wang, P., An, N., Cao, L., and Niu, Z.
\newblock Fourier{GNN}: Rethinking multivariate time series forecasting from a pure graph perspective.
\newblock In \emph{Thirty-seventh Conference on Neural Information Processing Systems}, 2023{\natexlab{a}}.
\newblock URL \url{https://openreview.net/forum?id=bGs1qWQ1Fx}.

\bibitem[Yi et~al.(2023{\natexlab{b}})Yi, Zhang, Fan, Wang, Wang, He, An, Lian, Cao, and Niu]{yi2023fmlp}
Yi, K., Zhang, Q., Fan, W., Wang, S., Wang, P., He, H., An, N., Lian, D., Cao, L., and Niu, Z.
\newblock Frequency-domain {MLP}s are more effective learners in time series forecasting.
\newblock In \emph{Thirty-seventh Conference on Neural Information Processing Systems}, 2023{\natexlab{b}}.
\newblock URL \url{https://openreview.net/forum?id=iif9mGCTfy}.

\bibitem[Zeng et~al.(2023)Zeng, Chen, Zhang, and Xu]{zeng2023dlinear}
Zeng, A., Chen, M., Zhang, L., and Xu, Q.
\newblock Are transformers effective for time series forecasting?
\newblock In \emph{Proceedings of the AAAI conference on artificial intelligence}, volume~37, pp.\  11121--11128, 2023.
\newblock URL \url{https://ojs.aaai.org/index.php/AAAI/article/view/26317/26089}.

\bibitem[Zhang et~al.(2022)Zhang, Zhang, Cao, Bian, Yi, Zheng, and Li]{Zhang2022lightts}
Zhang, T., Zhang, Y., Cao, W., Bian, J., Yi, X., Zheng, S., and Li, J.
\newblock Less is more: Fast multivariate time series forecasting with light sampling-oriented mlp structures.
\newblock \emph{arXiv preprint arXiv:2207.01186}, 2022.
\newblock URL \url{https://arxiv.org/abs/2207.01186}.

\bibitem[Zhang \& Yan(2023)Zhang and Yan]{zhang2023crossformer}
Zhang, Y. and Yan, J.
\newblock Crossformer: Transformer utilizing cross-dimension dependency for multivariate time series forecasting.
\newblock In \emph{The Eleventh International Conference on Learning Representations}, 2023.
\newblock URL \url{https://openreview.net/forum?id=vSVLM2j9eie}.

\bibitem[Zhao et~al.(2017)Zhao, Chen, Wu, Chen, and Liu]{Zhao2017traflstm}
Zhao, Z., Chen, W., Wu, X., Chen, P. C.~Y., and Liu, J.
\newblock Lstm network: a deep learning approach for short‐term traffic forecast.
\newblock \emph{IET Intelligent Transport Systems}, pp.\  68–75, Mar 2017.
\newblock \doi{10.1049/iet-its.2016.0208}.
\newblock URL \url{http://dx.doi.org/10.1049/iet-its.2016.0208}.

\bibitem[Zhou et~al.(2022{\natexlab{a}})Zhou, Zhang, Peng, Zhang, Li, Xiong, and Zhang]{Zhou2020informer}
Zhou, H., Zhang, S., Peng, J., Zhang, S., Li, J., Xiong, H., and Zhang, W.
\newblock Informer: Beyond efficient transformer for long sequence time-series forecasting.
\newblock \emph{Proceedings of the AAAI Conference on Artificial Intelligence}, pp.\  11106–11115, Sep 2022{\natexlab{a}}.
\newblock \doi{10.1609/aaai.v35i12.17325}.
\newblock URL \url{http://dx.doi.org/10.1609/aaai.v35i12.17325}.

\bibitem[Zhou et~al.(2022{\natexlab{b}})Zhou, MA, wang, Wen, Sun, Yao, Yin, and Jin]{Zhou2022film}
Zhou, T., MA, Z., wang, x., Wen, Q., Sun, L., Yao, T., Yin, W., and Jin, R.
\newblock Film: Frequency improved legendre memory model for long-term time series forecasting.
\newblock In Koyejo, S., Mohamed, S., Agarwal, A., Belgrave, D., Cho, K., and Oh, A. (eds.), \emph{Advances in Neural Information Processing Systems}, volume~35, pp.\  12677--12690. Curran Associates, Inc., 2022{\natexlab{b}}.

\bibitem[Zhou et~al.(2022{\natexlab{c}})Zhou, Ma, Wen, Wang, Sun, and Jin]{Zhou2022fedformer}
Zhou, T., Ma, Z., Wen, Q., Wang, X., Sun, L., and Jin, R.
\newblock Fedformer: Frequency enhanced decomposed transformer for long-term series forecasting.
\newblock \emph{CoRR}, abs/2201.12740, 2022{\natexlab{c}}.
\newblock URL \url{https://arxiv.org/abs/2201.12740}.

\end{thebibliography}
\bibliographystyle{icml2024}
\clearpage
\newpage
\appendix
\onecolumn
\section{Experimental Details}
\label{app:exp}
\subsection{Dataset Statistics}
\label{app:datasets details}
We elaborate on the datasets employed in this study with the following details.

\begin{itemize}
    \item \textbf{ETT Dataset}~\citep{Zhou2020informer} comprises two sub-datasets: \textbf{ETTh} and \textbf{ETTm}, which were collected from electricity transformers. Data were recorded at 15-minute and 1-hour intervals for ETTm and ETTh, respectively, spanning from July 2016 to July 2018.

    \item \textbf{Solar-Energy}~\citep{Lai2018lstnet} records the solar power production of 137 PV plants in 2006, which are sampled every 10 minutes.

    \item \textbf{Electricity1 Dataset}\footnote{\url{https://archive.ics.uci.edu/ml/datasets/ElectricityLoadDiagrams20112014}} encompasses the electricity consumption data of 321 customers, recorded on an hourly basis, covering the period from 2012 to 2014. 

    \item \textbf{Traffic Dataset}\footnote{\url{https://pems.dot.ca.gov/}} consists of hourly data from the California Department of Transportation. It describes road occupancy rates measured by various sensors on San Francisco Bay area freeways. 

    \item \textbf{Weather Dataset}\footnote{\url{https://www.bgc-jena.mpg.de/wetter/}} contains records of $21$ meteorological indicators, updated every $10$ minutes throughout the entire year of 2020. 

\end{itemize}
We follow the same data processing and train-validation-test set split protocol used in TimesNet~\citep{wu2022timesnet}, where the train, validation, and test datasets are strictly divided according to chronological order to make sure there are no data leakage issues. We fix the length of the lookback series as $T = 96$ for all datasets, and the prediction length $F \in \{96, 192, 336, 720\}$.

\subsection{Implementation Details and Model Parameters}
\label{app:implementation}
We trained our Leddam model using the L2 loss function and employed the ADAM optimizer. We initialized the random seed as $rs = 2021$. We also configured the hyperparameter $k = 25$-kernel size of the decomposition kernel (the Moving Average Kernel (MOV) and Learnable Decomposition Module (LD)). During the training process, we incorporated an early stopping mechanism, which would halt training after six epochs if no significant reduction in loss was observed on the validation set. For evaluation purposes, we used two key performance metrics: the mean square error (MSE) and the mean absolute error (MAE). We carried a grid hyperparameter search, where dimension of layer $dim \in \{ 256, 512\}$, learning rate $lr \ in \{0.001,0.0001, 0.0005\}$, dropout ratio $ dr \in \{0.0, 0.2, 0.5\}$ and number of network layers $nl \in \{1,2,3\}$. Our implementation was carried out in PyTorch and executed on an NVIDIA V100 32GB GPU. All the compared baseline models that we reproduced are implemented based on the benchmark of TimesNet~\citep{wu2022timesnet} Repository, which is fairly built on the configurations provided by each model's original paper or official code. It is worth noting that both the baselines used in this paper and our Leddam have fixed a long-standing bug. This bug was originally identified in Informer~\citep{Zhou2020informer} \textbf{(AAAI 2021 Best Paper)} and subsequently addressed by FITS~\citep{anonymous2024fits}. For specific details about the bug and its resolution, please refer to \textbf{GitHub Repository}\footnote{\url{https://github.com/VEWOXIC/FITS}}.

\section{Further Analysis of Different Decomposition Methodologies}

\subsection{Decomposition Result Visualization}
\label{app:decom results visualization}
\begin{figure*}[t]
   \centering
   \includegraphics[width=1.0\textwidth]{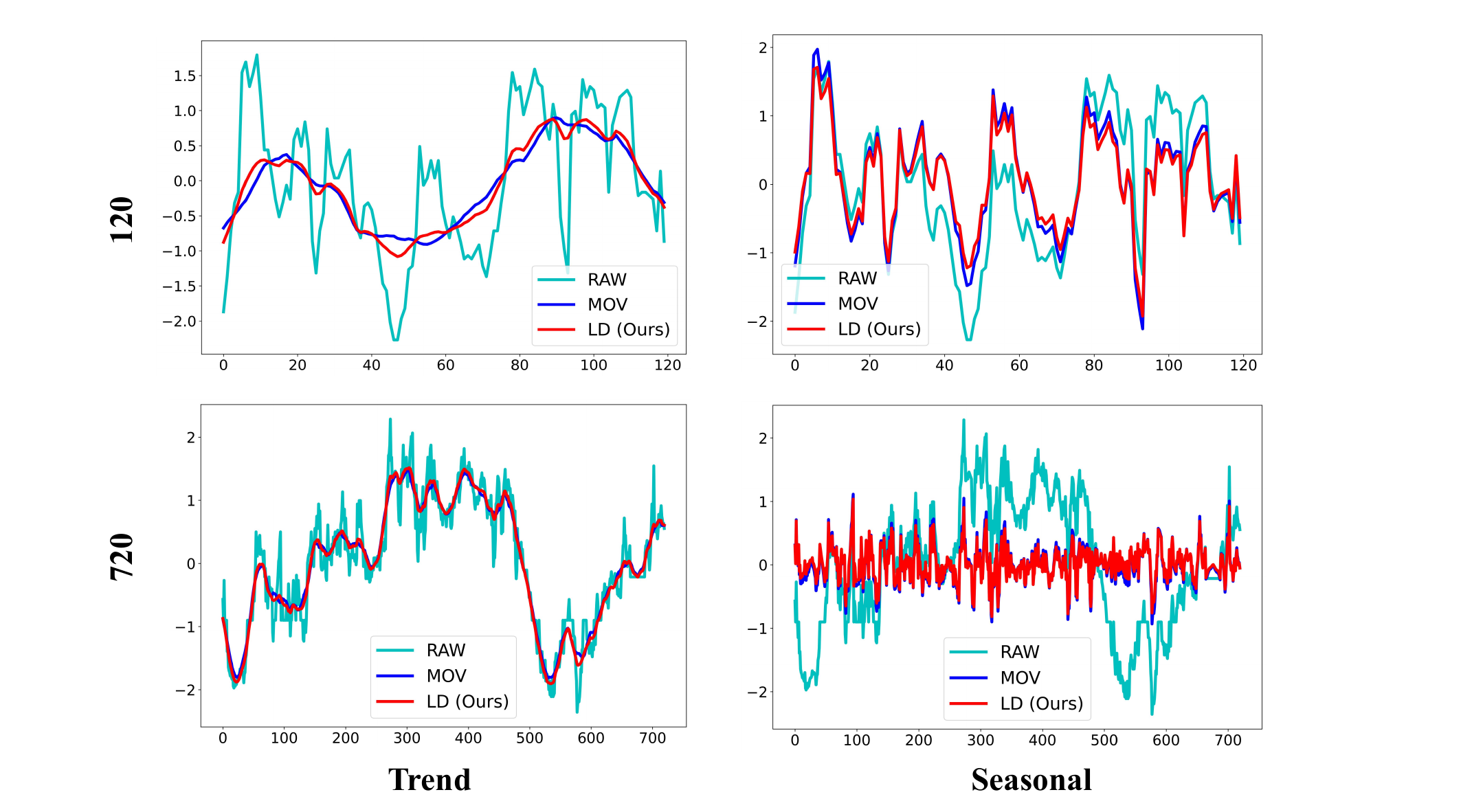}
   \caption{Trend-Seasonal Decomposition Results obtained by LD (Red) and MOV (Blue) on ETTh1.}
   \label{figure:decom_1}
\end{figure*}
\begin{figure*}[t]
   \centering
   \includegraphics[width=1.0\textwidth]{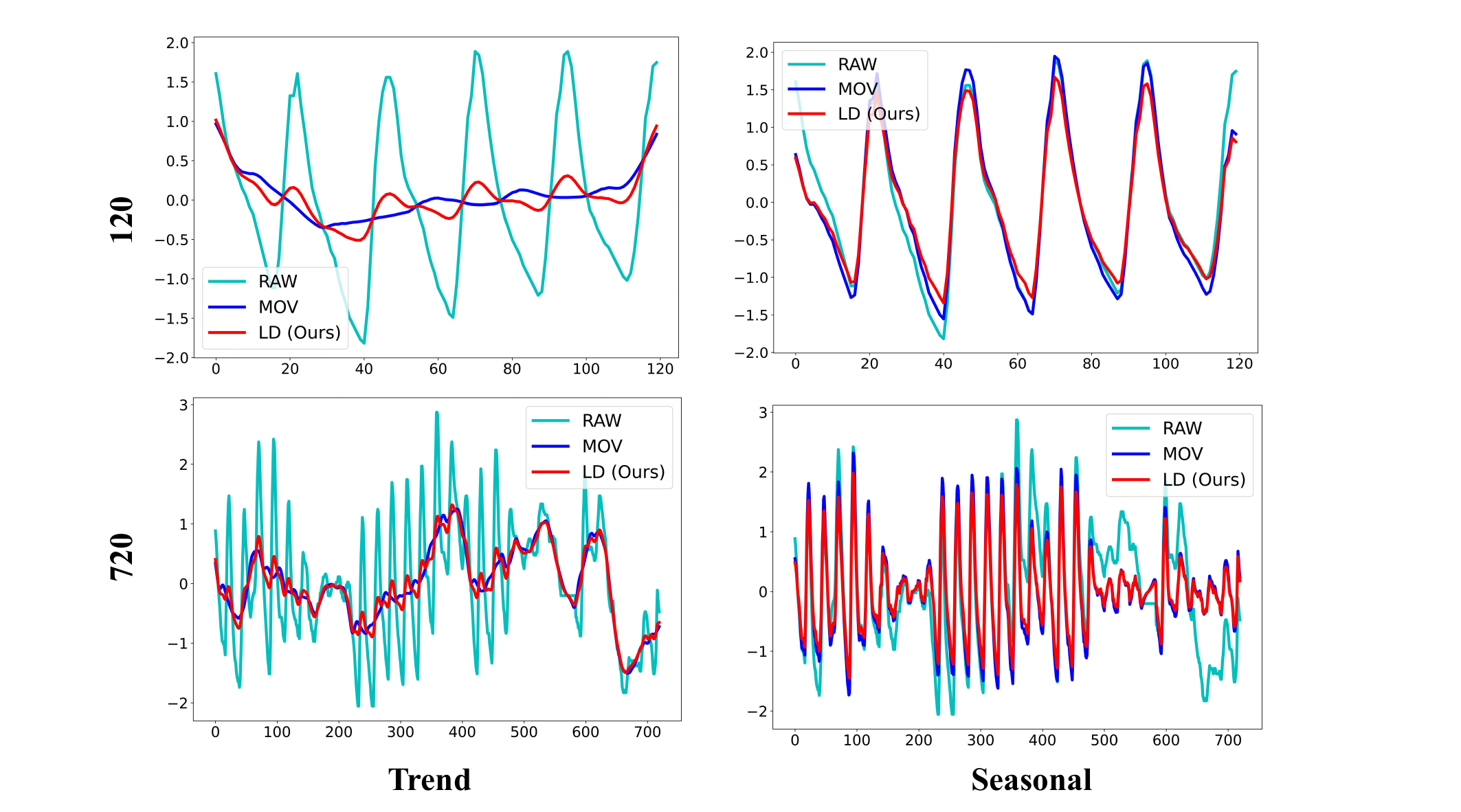}
   \caption{Trend-Seasonal Decomposition Results obtained by LD (Red) and MOV (Blue) on ETTh2.}
   \label{figure:decom_2}
\end{figure*}
\begin{figure*}[t]
   \centering
   \includegraphics[width=1.0\textwidth]{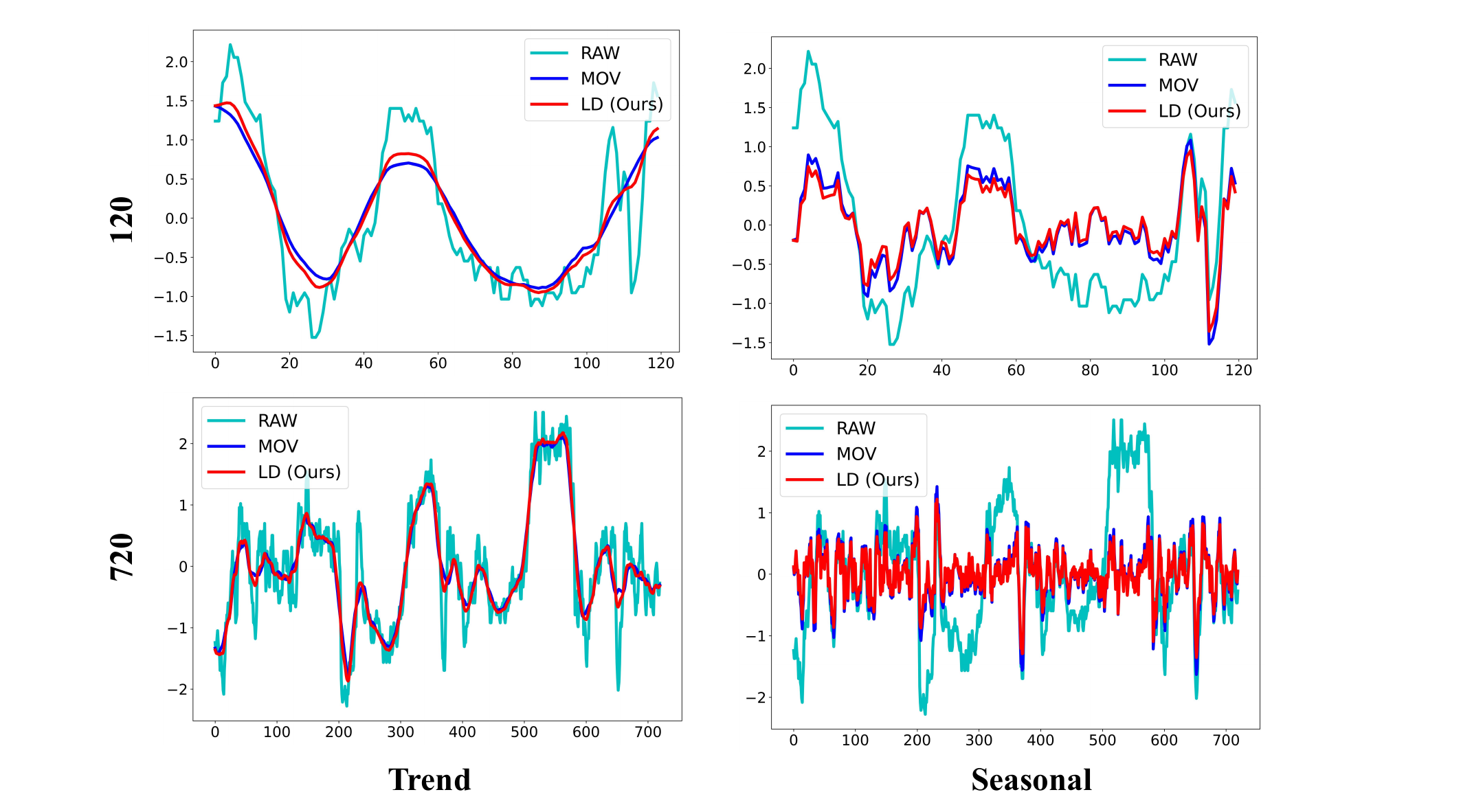}
   \caption{Trend-Seasonal Decomposition Results obtained by LD (Red) and MOV (Blue) on ETTm1.}
   \label{figure:decom_3}
\end{figure*}
\begin{figure*}[t]
   \centering
   \includegraphics[width=1.0\textwidth]{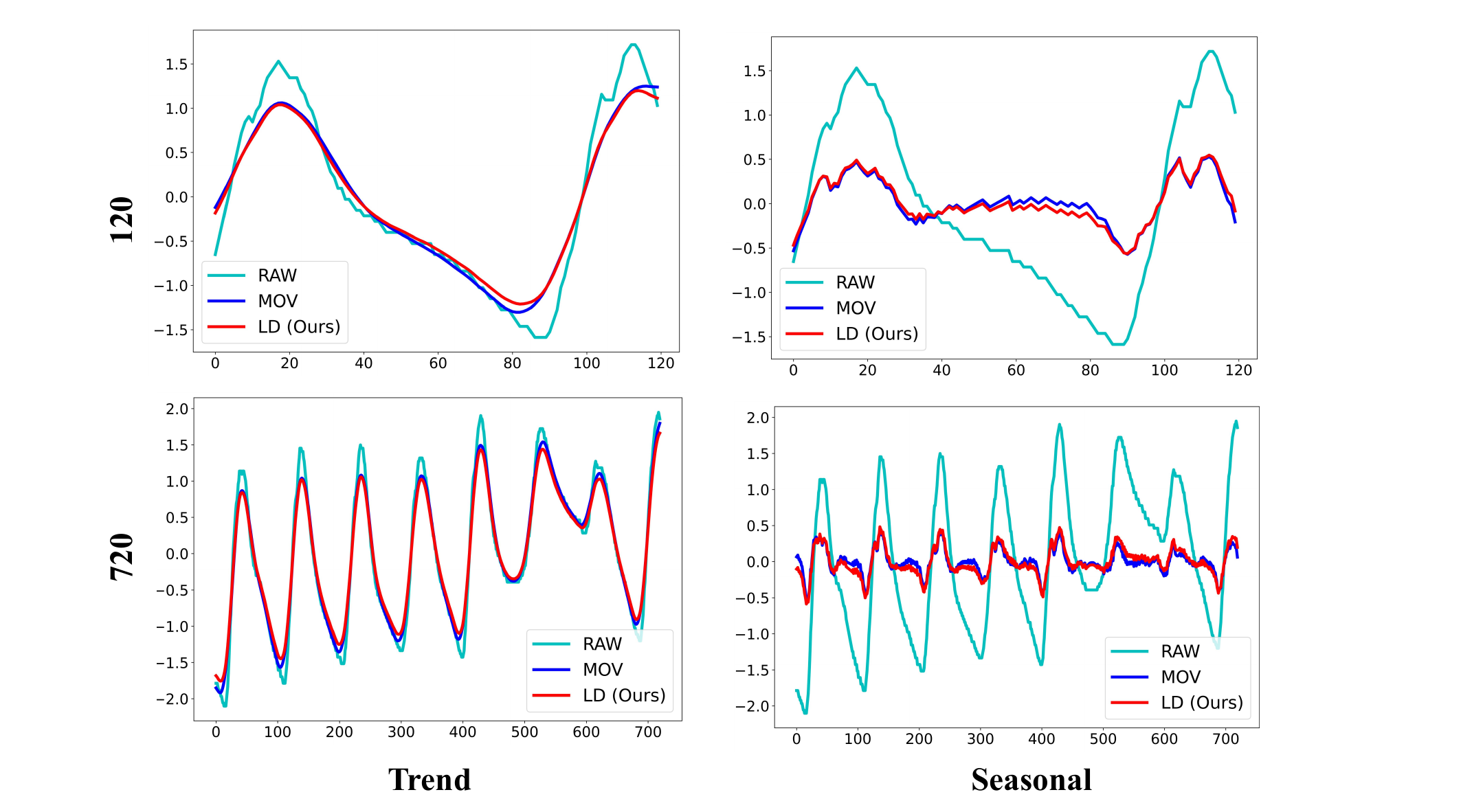}
   \caption{Trend-Seasonal Decomposition Results obtained by LD (Red) and MOV (Blue) on ETTm2.}
   \label{figure:decom_4}
\end{figure*}
\begin{figure*}[t]
   \centering
   \includegraphics[width=1.0\textwidth]{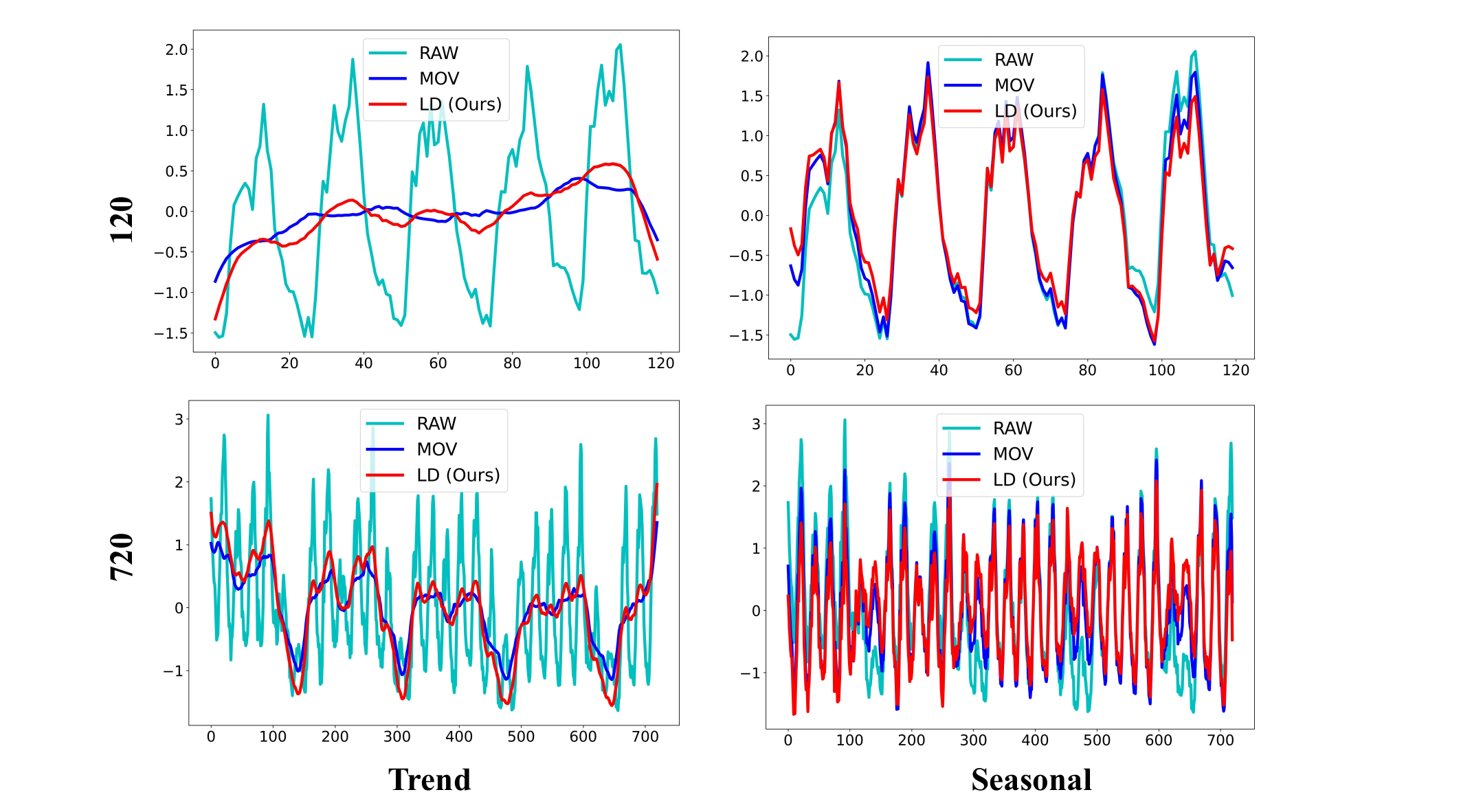}
   \caption{Trend-Seasonal Decomposition Results obtained by LD (Red) and MOV (Blue) on Electricity.}
   \label{figure:decom_5}
\end{figure*}
\begin{figure*}[t]
   \centering
   \includegraphics[width=1.0\textwidth]{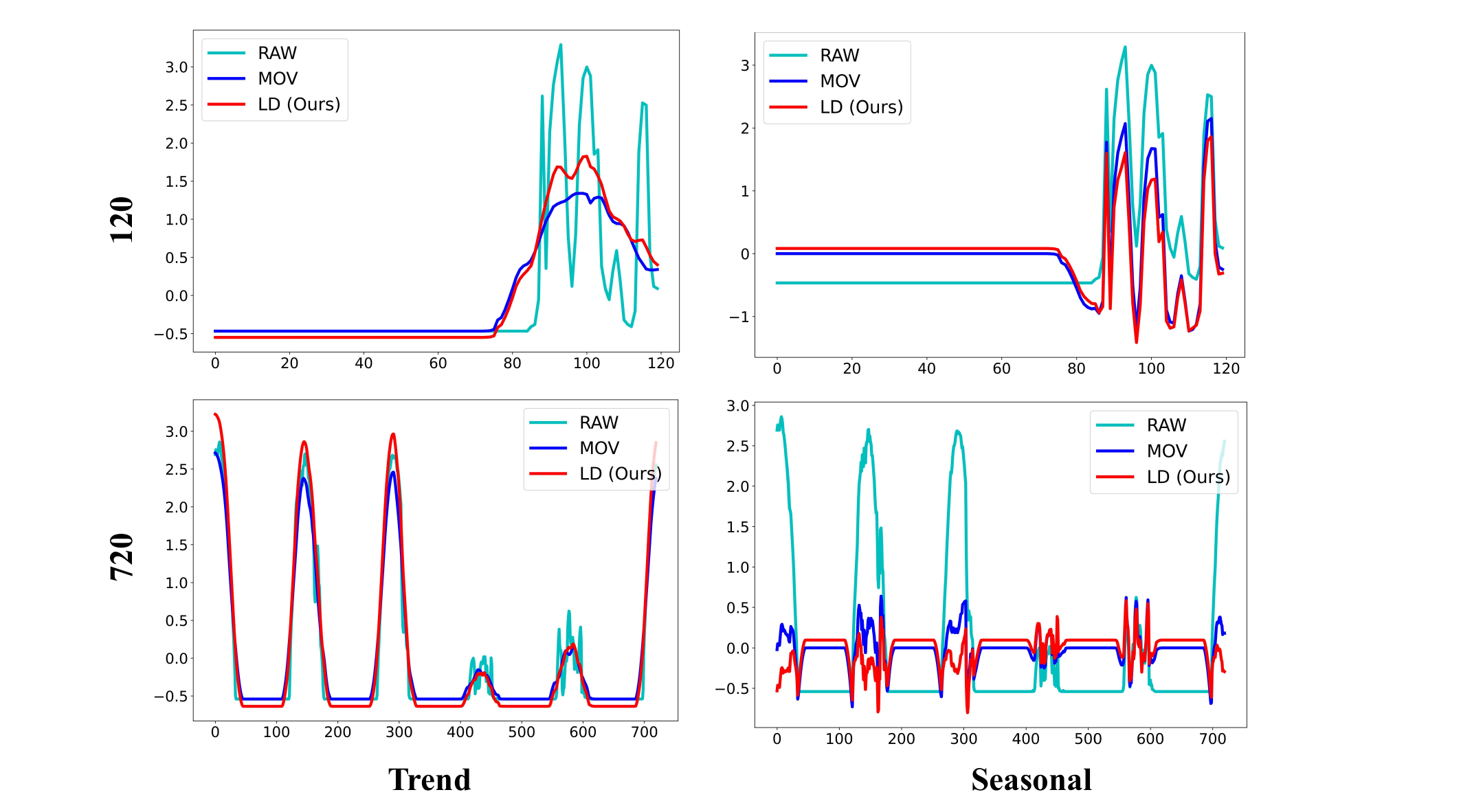}
   \caption{Trend-Seasonal Decomposition Results obtained by LD (Red) and MOV (Blue) on Solar-Energy.}
   \label{figure:decom_6}
\end{figure*}
\begin{figure*}[t]
   \centering
   \includegraphics[width=1.0\textwidth]{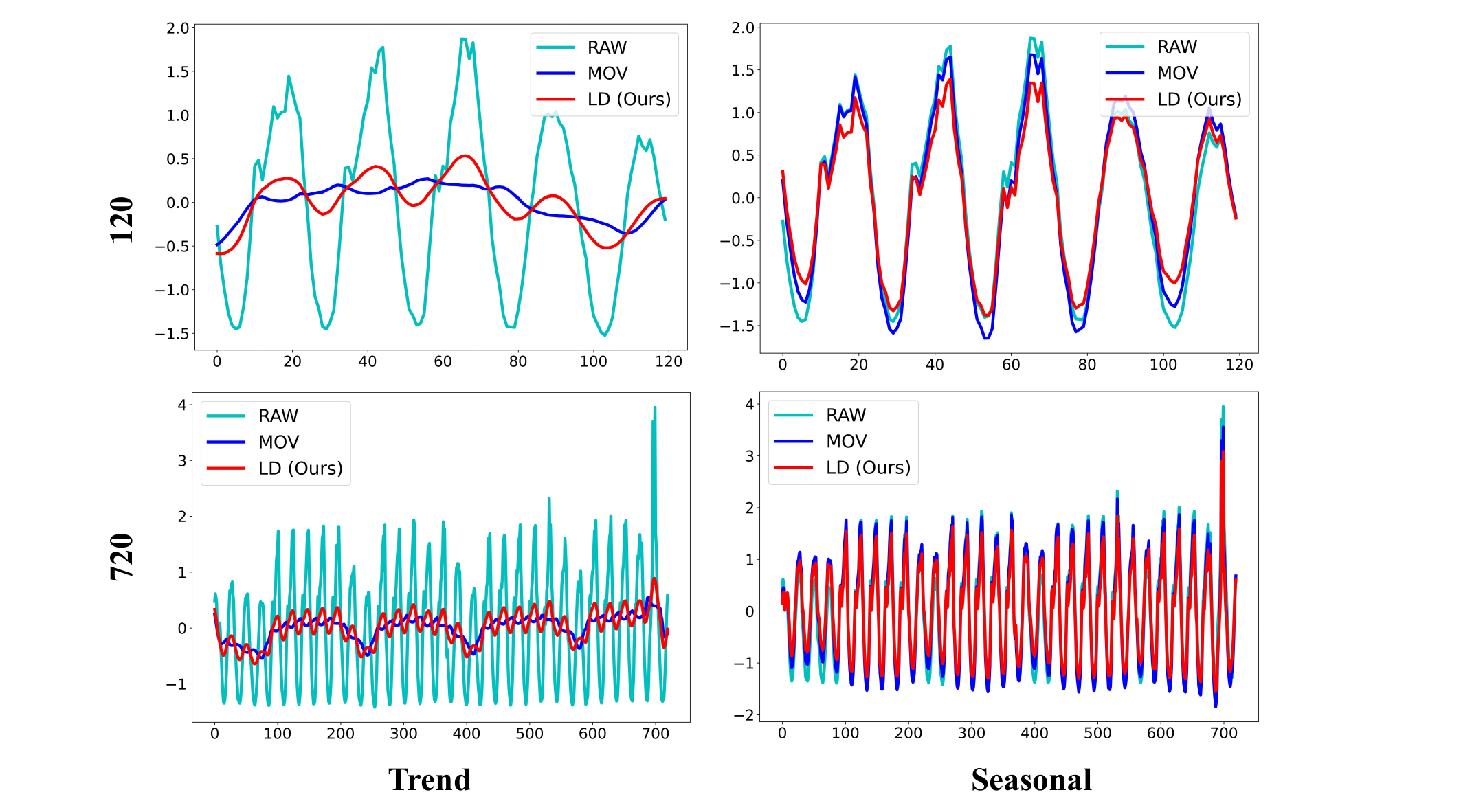}
   \caption{Trend-Seasonal Decomposition Results obtained by LD (Red) and MOV (Blue) on Traffic.}
   \label{figure:decom_7}
\end{figure*}
\begin{figure*}[t]
   \centering
   \includegraphics[width=1.0\textwidth]{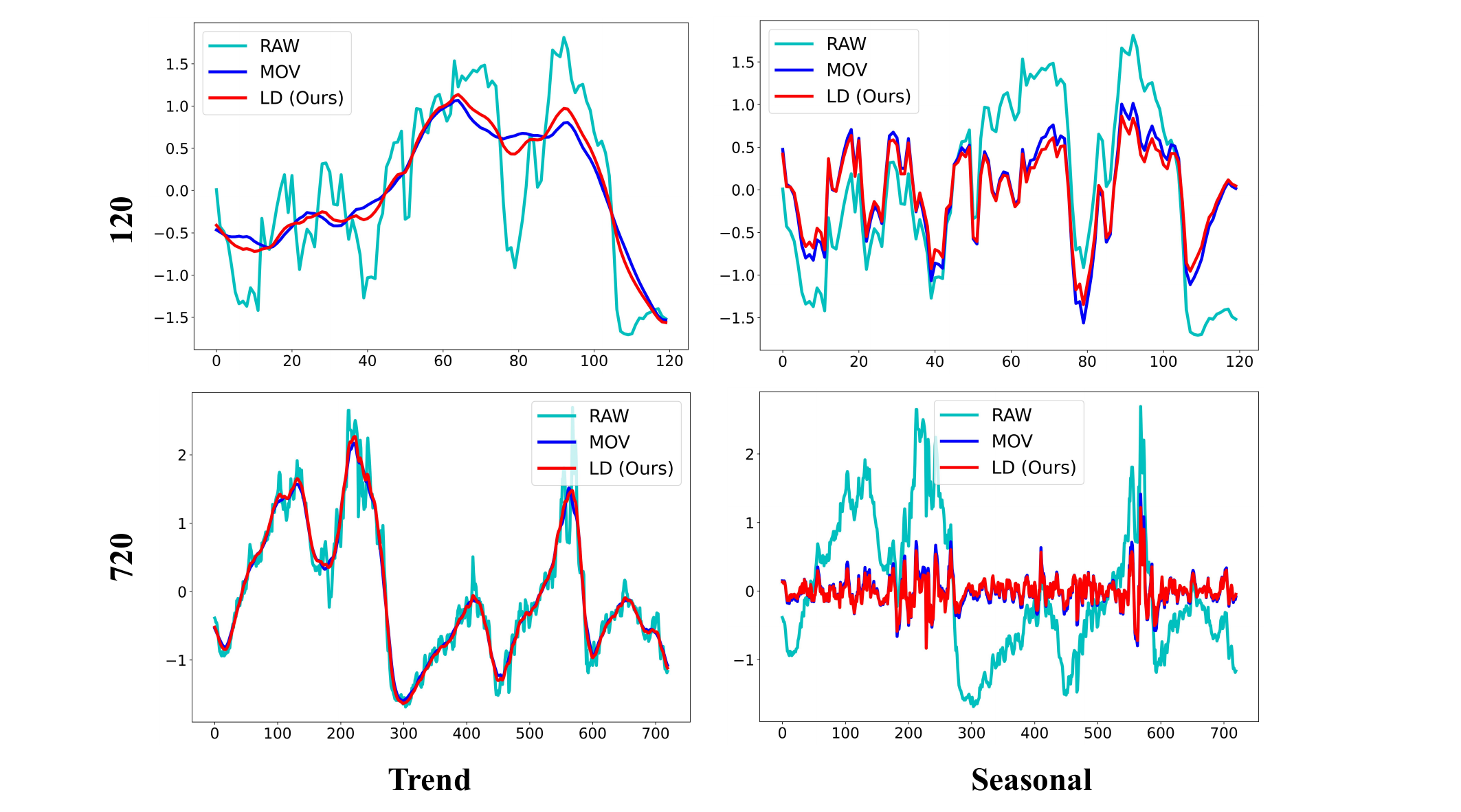}
   \caption{Trend-Seasonal Decomposition Results obtained by LD (Red) and MOV (Blue) on Weather.}
   \label{figure:decom_8}
\end{figure*}
We present a detailed comparison of results obtained by decomposing various datasets using the Moving Average Kernel (MOV) and Learnable Decomposition Module (LD). 

We still selected \textbf{the final variate of each dataset}. LD are pretrained on task: input-96-forecast-720. Given a sampling frequency of 10, 15 minutes, or 1 hour for these datasets, we opt to decompose time series of two lengths: 120 and 720, to better contrast the decomposition results and reflect the extracted seasonal and trend patterns respectively.

The decomposition performance of LD is consistently superior to that of the MOV across all eight datasets in Figure~\ref{figure:decom_1}–\ref{figure:decom_8}.

\subsection{Visualization of Weights}
\label{app:weight visualization}
\begin{figure*}[t]
   \centering
   \includegraphics[width=1.0\textwidth]{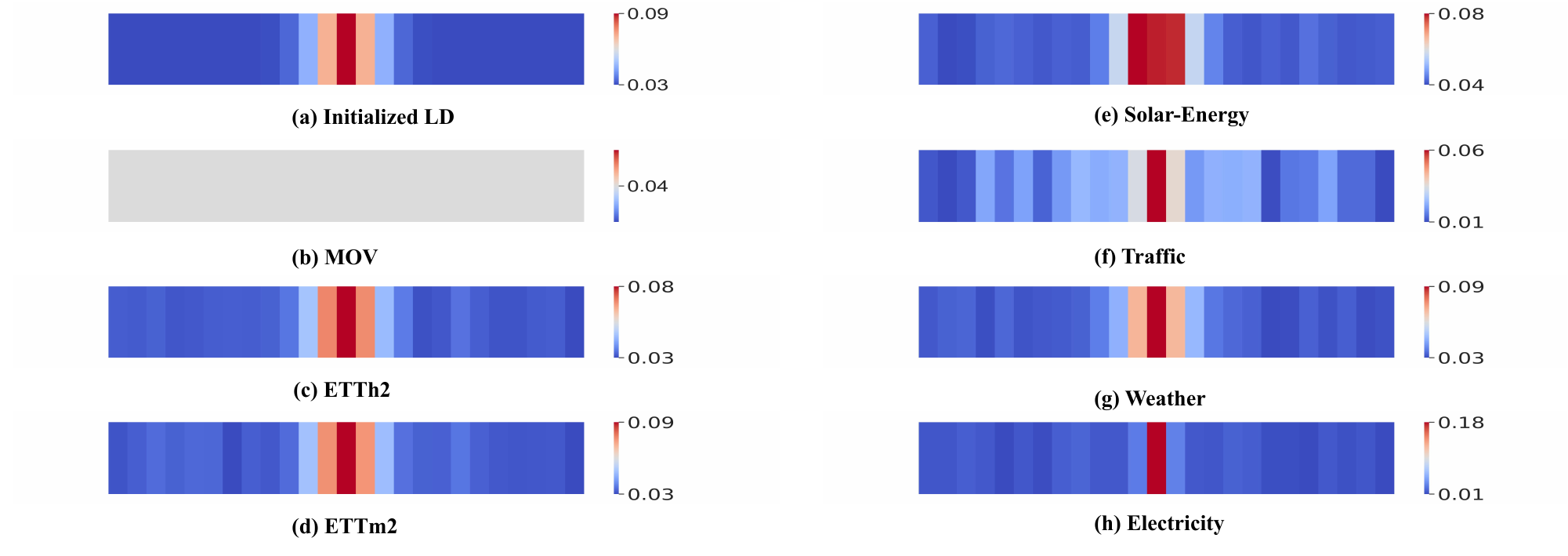}
   \caption{Weight Visualization of the Learnable Decomposition Module (LD) and Moving Average Kernel (MOV). (a) The weight of initialized LD and (b) MOV are the same for all datasets. (c)-(h) is the weight visualization of the LD after training on other datasets.}
   \label{figure:wv}
\end{figure*}
We present visualizations of the weight for the Learnable Decomposition Module (LD) and Moving Average Kernel (MOV) in Figure~\ref{figure:wv}. It is observed that, unlike MOV, which employs uniform weights across all datasets, our LD adapts to the characteristics of different datasets, generating context-specific weights for sequence decomposition.

\subsection{Impact of RevIN}
\label{app:RevIN}
To explore how much ReVIN~\citep{Kim_revin} improves the performance of LD (and MOV), we utilized Autoformer~\citep{Wu2021autoformer} as the baseline and investigated the performance of LD and MOV under scenarios with and without ReVIN. In Table~\ref{t:revin}, We computed that ReVIN yielded an average MSE performance improvement (reduction) of 5.24\% across all tasks for LD, and 9.04\% for MAE. For MOV: 1.75\% for MSE and 3.32\% for MAE. Compared to MOV, ReVIN would bring greater performance improvements to LD.

\begin{table*}[t]
\centering
\caption{Model Comparison}
\label{t:revin}
\resizebox{0.9\textwidth}{!}{
\begin{tabular}{c|c|cc|cc|cc|cc}
\toprule
  \multirow{2}{*}{Dataset/Metric}  & \multirow{2}{*}{pred\_len} & \multicolumn{2}{c}{LD+ReVIN} & \multicolumn{2}{c}{MOV+ReVIN} & \multicolumn{2}{c}{LD} & \multicolumn{2}{c}{MOV} \\
   \cmidrule{3-10}
& & MSE & MAE & MSE & MAE & MSE & MAE & MSE & MAE \\
\midrule
\multirow{2}{*}{ETTm1} & 96 & 0.667 & 0.526 & 0.694 & 0.538 & 0.664 & 0.531 & 0.671 & 0.534 \\
 & 720 & 0.659 & 0.538 & 0.753 & 0.559 & 0.647 & 0.547 & 0.688 & 0.556 \\
 \cmidrule{1-10}
\multirow{2}{*}{ETTm2} & 96 & 0.207 & 0.281 & 0.218 & 0.292 & 0.216 & 0.299 & 0.223 & 0.306 \\
 & 720 & 0.414 & 0.404 & 0.422 & 0.41 & 0.435 & 0.43 & 0.426 & 0.423 \\
  \cmidrule{1-10}
\multirow{2}{*}{Weather} & 96 & 0.201 & 0.249 & 0.217 & 0.266 & 0.289 & 0.366 & 0.233 & 0.304 \\
 & 720 & 0.373 & 0.369 & 0.378 & 0.373 & 0.402 & 0.408 & 0.394 & 0.396 \\
\bottomrule
\end{tabular}
}
\end{table*}

\subsection{Further Improvement of Multi-scale Hybrid Decomposition}
\label{app:ms_deom}
In MICN~\citep{wang2023micn}, it is observed that they employ a multi-scale decomposition strategy akin to FEDformer~\citep{Zhou2022fedformer}. Specifically, they utilize decomposition kernels of varying scales, followed by integrating all decomposed outputs to derive the trend part:
\begin{align}
    i& =1,\ldots,n \notag \\
    X_{Trend}^{i} &= AvgPool( Padding(X_{embed}))_{kernel_{i}} \notag \\
    X_{Trend} &= Intergrating(X_{Trend}^{1},\ldots,X_{Trend}^{n}) \notag \\
    X_{Seasonal} &= X_{embed}-X_{Trend}
\end{align}
MICN argues that using simple mean operations to integrate these different patterns is a superior plan. 

We also conducted exploratory experiments to investigate the performance of our Learnable Decomposition Module (LD) in this particular context. We replaced the fundamental decomposition kernel employed in MICN, i.e., the basic Moving Average Kernel (MOV), with our LD. Subsequently, comparative experiments were conducted on four datasets(ETTm2, Weather, Electricity, Traffic) with input length $T = 96$ and prediction length $F = 96$. The LD consistently outperforms MOV under a multi-scale decomposition strategy across all datasets in Table~\ref{t:ms_deom}, which highlights the excellence of LD.
\begin{table*}[t]
\centering
\caption{Comparision of Learnable Decomposition Module (LD) and Moving Average Kernel (MOV) under multi-scale decomposition strategy across ETTm2, Weather, Electricity, and Traffic datasets with prediction lengths $F = 96$, and input length $T = 96$.}
\label{t:ms_deom}
\resizebox{0.7\textwidth}{!}{
\begin{tabular}{ccccccccc}
\toprule
\multirow{2}{*}{Design} & \multicolumn{2}{c}{ETTm2} & \multicolumn{2}{c}{Electricity} & \multicolumn{2}{c}{Traffic} & \multicolumn{2}{c}{Weather}\\
\cmidrule(lr){2-3} \cmidrule(lr){4-5} \cmidrule(lr){6-7} \cmidrule(lr){8-9} 
& MSE & MAE & MSE & MAE & MSE & MAE & MSE & MAE \\
\midrule
Original & 0.197 & 0.296  & 0.202  & 0.268  & 0.172  & 0.279  & 0.530  & 0.321 \\
\cmidrule(lr){1-9}
\textbf{LD} &        \textbf{0.183} & \textbf{0.281}  & \textbf{0.170}  & \textbf{0.230}  & \textbf{0.164}  & \textbf{0.276} & \textbf{0.515}  & \textbf{0.310} \\
\cmidrule(lr){1-9}
 \textbf{Decrease}  & \textcolor{red}{\textbf{6.87\%}}  & \textcolor{red}{\textbf{5.07\%}}   & \textcolor{red}{\textbf{15.97\%}}  & \textcolor{red}{\textbf{14.02\%}}  & \textcolor{red}{\textbf{4.71\%}}  & \textcolor{red}{\textbf{1.11\%}}  & \textcolor{red}{\textbf{2.83\%}}  & \textcolor{red}{\textbf{3.28\%}} \\
\bottomrule
\end{tabular}
}
\end{table*}

\section{Further Analysis of Auto-regressive Self-attention}
\label{app:auto}
Our experiments in Table~\ref{t:embed_comparison} demonstrate that our Auto-regressive Embedding, compared to Point-wise and Patch-wise Embedding, maintains crucial temporal positional information within permutation-invariant self-attention mechanisms. This is demonstrated by the low dependency of Auto-regressive Embedding on positional encoding.
\begin{table*}[t]
\centering
\caption{Forecasting performance of Auto-regressive, Point-wise, and Patch-wise Embedding on the ETTh2, ETTm2, and Weather datasets. The forecasting task is input-96-forecast-96/720. \textbf{w/o} means without position embedding, \textbf{pos} means position embedding is used.}
\label{t:embed_comparison}
\resizebox{0.7\textwidth}{!}{
\begin{tabular}{c|c|c|cc|cc|cc}
\toprule
Models & pred\_len & Type & \multicolumn{2}{c}{Auto} & \multicolumn{2}{c}{Patch} & \multicolumn{2}{c}{Point} \\
\midrule
Dataset/Metric & & & MSE & MAE & MSE & MAE & MSE & MAE \\
\midrule
\multirow{4}{*}{ETTh2} & \multirow{2}{*}{96} & w/o & 0.294 & 0.343 & 0.382 & 0.429 & 1.453 & 1.000 \\
 & & pos & 0.293 & 0.343 & 0.354 & 0.404 & 0.958 & 0.803 \\ \cmidrule{3-9}
 & \multirow{2}{*}{720} & w/o & 0.413 & 0.432 & 0.926 & 0.703 & 3.533 & 1.625 \\
 & & pos & 0.416 & 0.433 & 0.835 & 0.666 & 1.569 & 0.991 \\
\midrule
\multirow{4}{*}{ETTm2} & \multirow{2}{*}{96} & w/o & 0.176 & 0.257 & 0.226 & 0.321 & 0.419 & 0.511 \\
 & & pos & 0.175 & 0.256 & 0.208 & 0.314 & 0.335 & 0.447 \\
 \cmidrule{3-9}
 & \multirow{2}{*}{720} & w/o & 0.398 & 0.395 & 0.963 & 0.726 & 2.826 & 1.386 \\
 & & pos & 0.393 & 0.394 & 0.732 & 0.671 & 1.411 & 0.996 \\
\midrule
\multirow{4}{*}{Weather} & \multirow{2}{*}{96} & w/o & 0.166 & 0.211 & 0.176 & 0.240 & 0.219 & 0.309 \\
 & & pos & 0.167 & 0.213 & 0.165 & 0.234 & 0.174 & 0.260 \\
 \cmidrule{3-9}
 & \multirow{2}{*}{720} & w/o & 0.342 & 0.343 & 0.355 & 0.405 & 0.351 & 0.395 \\
 & & pos & 0.342 & 0.344 & 0.337 & 0.370 & 0.346 & 0.388 \\
\bottomrule
\end{tabular}
}
\end{table*}

\section{Generalization Analysis of Leddam}
\label{app:generalization}
\begin{figure*}[t]
   \centering
   \includegraphics[width=0.95\textwidth]{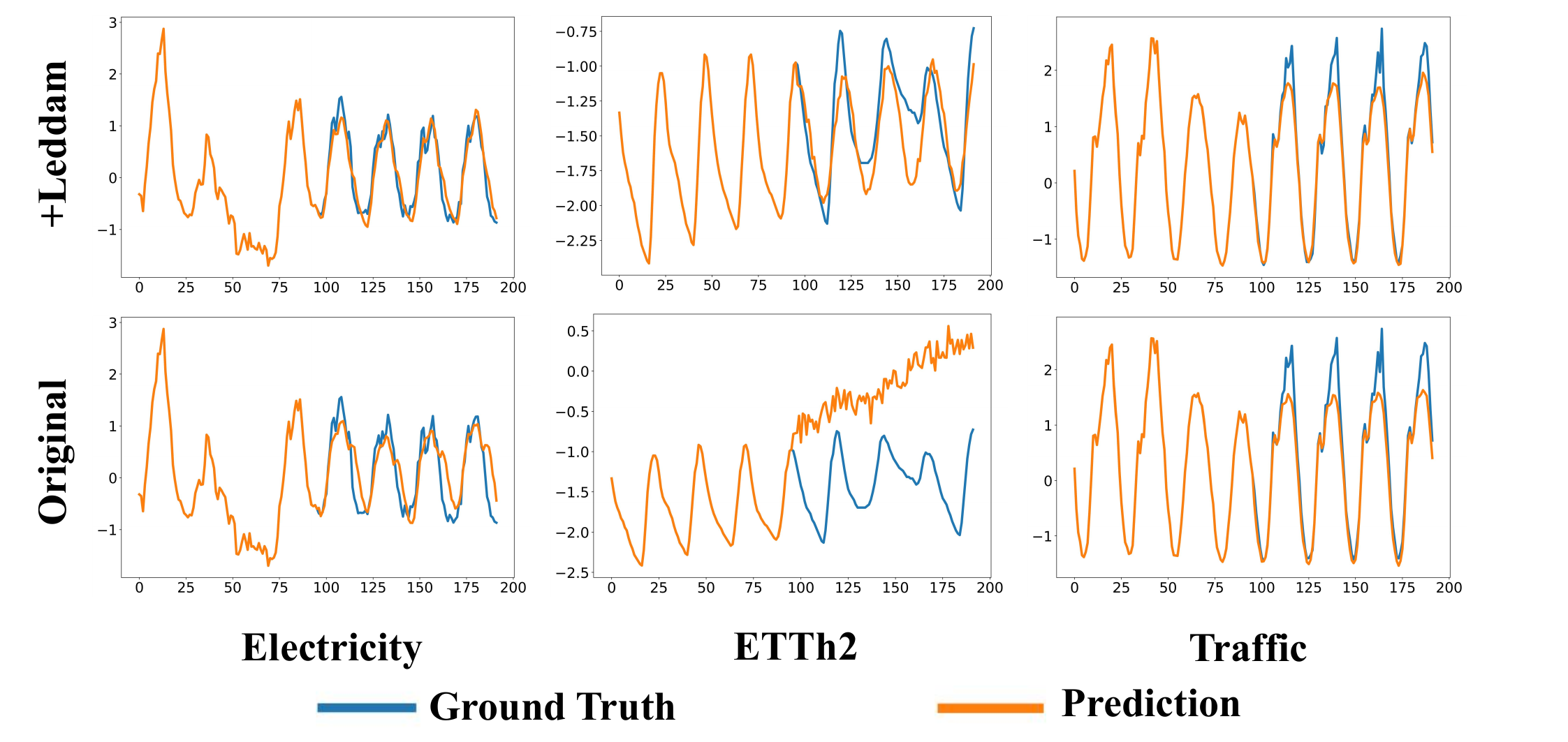}
   \caption{Visualization of input-96-predicts-96 results of LSTM (with and without Leddam) on three datasets (Electricity, ETTh2, Traffic). The above rows represent the performance incorporating our Leddam framework, while the below rows depict the predictive performance of the original model.}
   \label{figure:b_LSTM}
\end{figure*}

\begin{figure*}[t]
   \centering
   \includegraphics[width=0.95\textwidth]{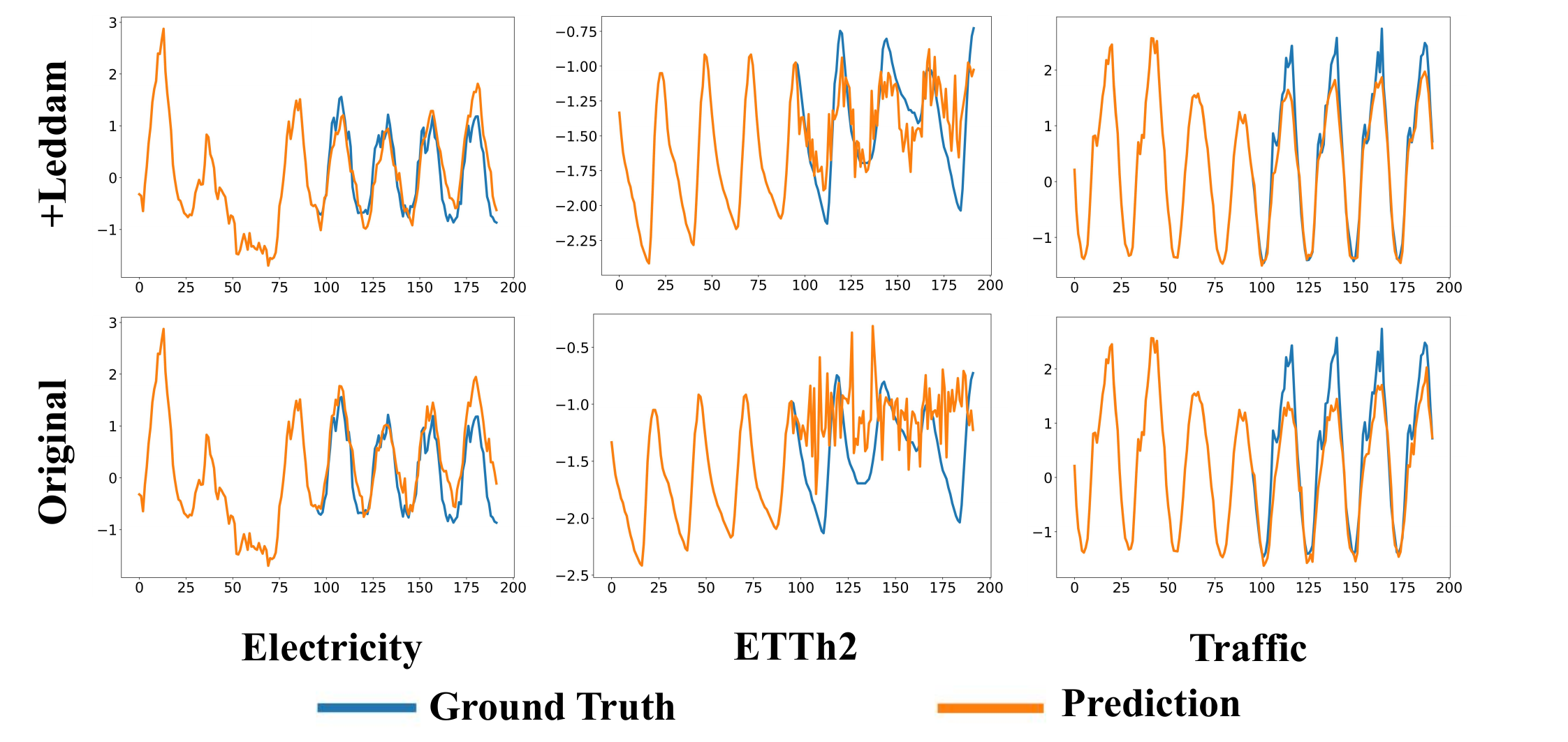}
   \caption{Visualization of input-96-predicts-96 results of SCINet (with and without Leddam) on three datasets (Electricity, ETTh2, Traffic). The above rows represent the performance incorporating our Leddam framework, while the rows below depict the predictive performance of the original model.}
   \label{figure:b_SCINet}
\end{figure*}

\begin{figure*}[t]
   \centering
   \includegraphics[width=0.95\textwidth]{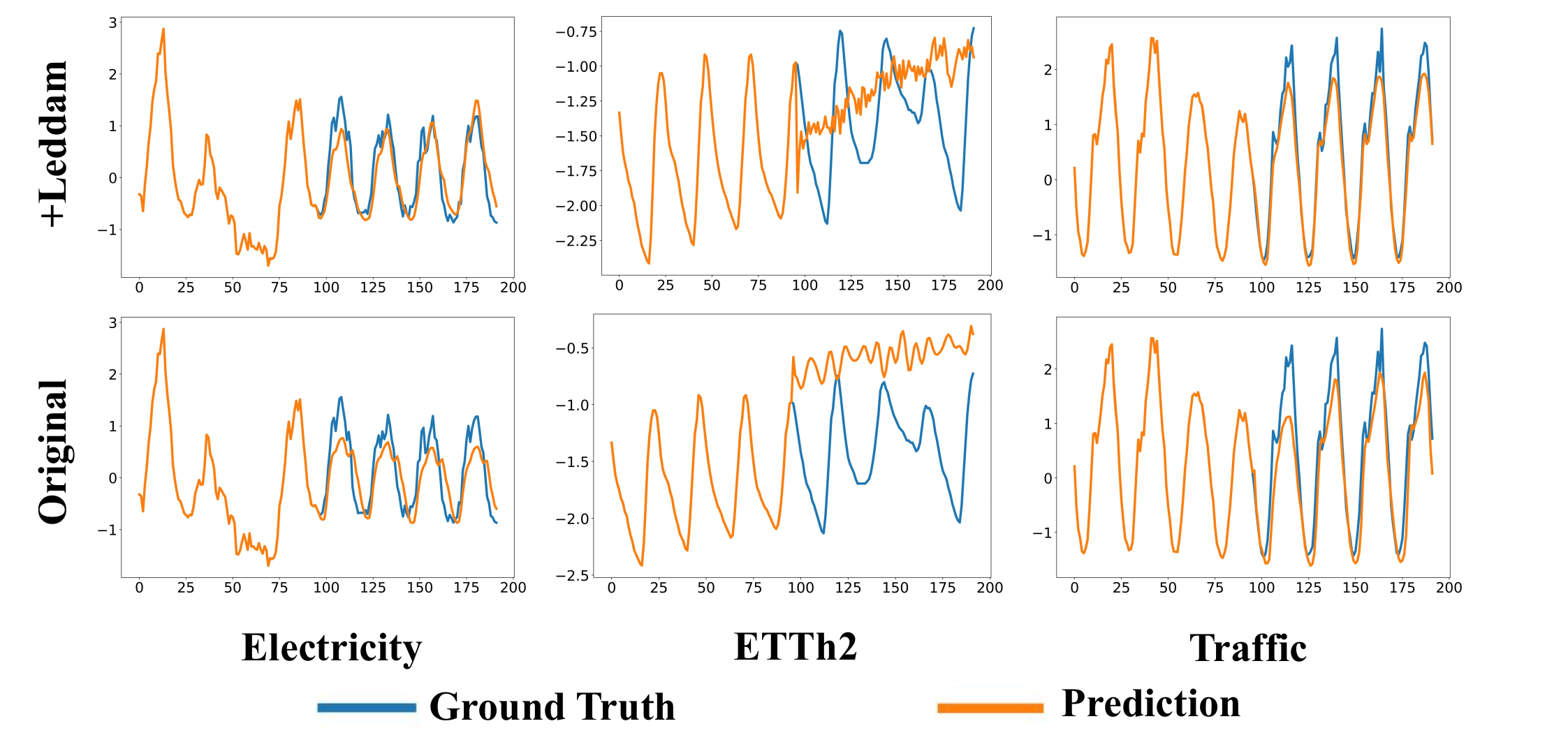}
   \caption{Visualization of input-96-predicts-96 results of Transformer (with and without Leddam) on three datasets (Electricity, ETTh2, Traffic). The above rows represent the performance incorporating our Leddam framework, while the rows below depict the predictive performance of the original model.}
   \label{figure:b_Transformer}
\end{figure*}

\begin{figure*}[t]
   \centering
   \includegraphics[width=0.95\textwidth]{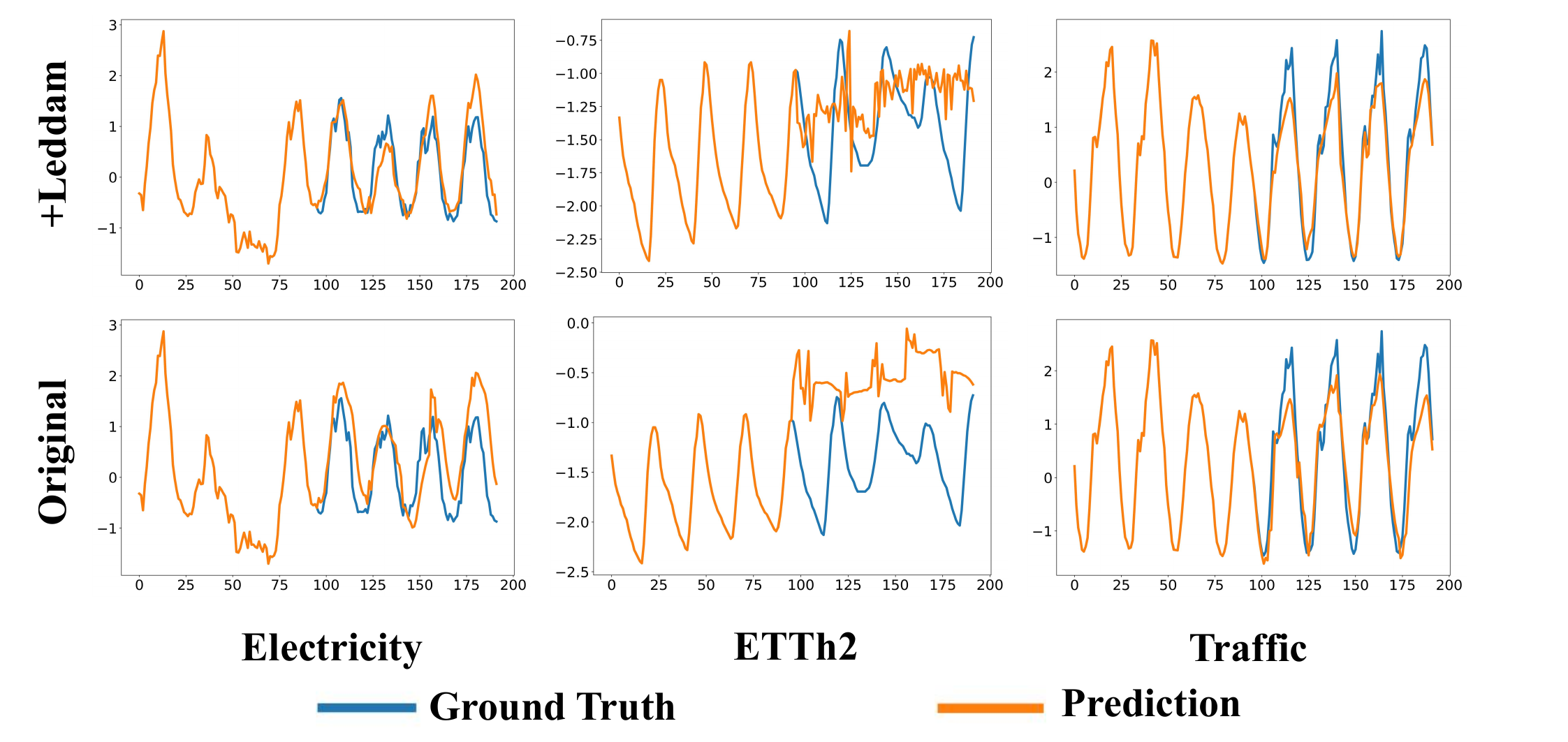}
   \caption{Visualization of input-96-predicts-96 results of Informer (with and without Leddam) on three datasets (Electricity, ETTh2, Traffic). The above rows represent the performance incorporating our Leddam framework, while the rows below depict the predictive performance of the original model.}
   \label{figure:b_Informer}
\end{figure*}

\begin{figure*}[t]
   \centering
   \includegraphics[width=0.95\textwidth]{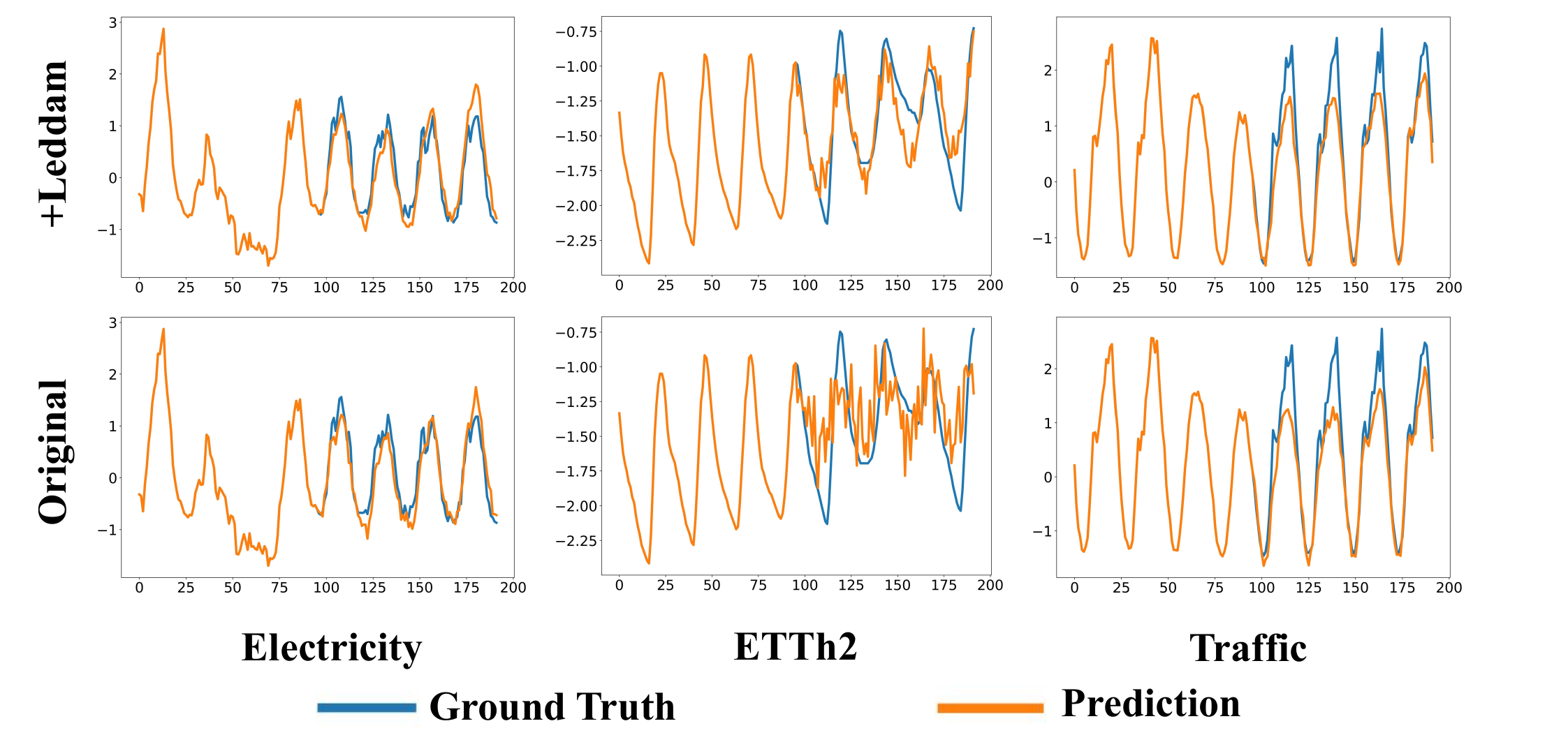}
   \caption{Visualization of input-96-predicts-96 results of LightTS (with and without Leddam) on three datasets (Electricity, ETTh2, Traffic). The above rows represent the performance incorporating our Leddam framework, while the rows below depict the predictive performance of the original model.}
   \label{figure:b_LightTS}
\end{figure*}

To better illustrate the generality of our Leddam framework and its performance improvement across various models, we opted to visualize the predictive outcomes on three representative datasets (Electricity, ETTh2, Traffic) in Figure~\ref{figure:b_LSTM}-\ref{figure:b_LightTS}. For a prediction task with input length $T = 96$ and prediction length $F = 96$, we present a comparative analysis of the predictive performance of different models before and after integrating our Leddam framework. To achieve this objective, we conduct experiments across a spectrum of representative time series forecasting model structures, including (1) Transformer-based methods: Informer~\citep{Zhou2020informer}, Transformer~\citep{Vaswani2017transformer}; (2) Linear-based methods: LightTS~\citep{Zhang2022lightts}; and (3) TCN-based methods: SCINet~\cite{Liu2021scinet}; (4) RNN-based methods: LSTM~\citep{Hochreiter1997lstm}. After the introduction of the Leddam framework, various models have consistently demonstrated superior predictive performance.

\section{Model Efficiency Study}
\label{app:efficiency}
We evaluated the \textbf{parameter count}, and the \textbf{inference time} (average of 5 runs on a single NVIDIA V100 32GB GPU) with $batch\_size= 1$ on \textbf{ETTh1} and \textbf{Electricity} dataset. We set the dimension of layer $ dim \in \{96, 192, 336, 720\}$, and the number of network layers $nl=2$. The task is input-96-forecast-720. We explored \textbf{Leddam} and four cutting-edge transformer-based multivariate time series forecasting models:\textbf{iTransformer}, \textbf{Crossformer}, \textbf{PatchTST}, and \textbf{FEDformer}. Results can be found in Table~\ref{t:eff_comparison}.

\begin{table*}[t]
\centering
\caption{Model efficiency analysis. \textbf{*} means `former.' \textbf{Para} means `Parameter count(M).' \textbf{Time} means `inference time(ms).'}
\label{t:eff_comparison}
\resizebox{0.9\textwidth}{!}{
\begin{tabular}{c|c|cc|cc|cc|cc|cc }
\toprule
\multirow{2}{*}{Datasets/Models} & \multirow{2}{*}{dim} & \multicolumn{2}{c}{Leddam} & \multicolumn{2}{c}{PatchTST} & \multicolumn{2}{c}{Cross*} & \multicolumn{2}{c}{iTrans*} & \multicolumn{2}{c}{FED*} \\
\cmidrule{3-12}
&  & Param & Time & Para & Time & Para & Time & Para & Time & Para & Time \\
\midrule
\multirow{2}{*}{ETTh1} & 256 & 2.50 & 233.92 & 3.27 & 251.00 & 8.19 & 399.00 & \textbf{1.27} & \textbf{177.67} & 3.43 & 303.556 \\
& 512 & 9.20 & 249.34 & 8.64 & 266.66 & 32.11 & 445.74 & \textbf{4.63} & \textbf{190.92} & 13.68 & 345.736 \\
\cmidrule{1-12}
\multirow{2}{*}{Electricity} & 256 & 2.59 & 283.04 & 3.27 & 322.53 & 13.66 & 432.40 & \textbf{1.27} & \textbf{192.12} & 4.24 & 347.634 \\
& 512 & 9.36 & 296.70 & 8.64 & 411.96 & 43.04 & 507.54 & \textbf{4.63} & \textbf{249.60} & 15.29 & 398.599 \\
\bottomrule
\end{tabular}
}
\end{table*}

\section{Hyperparameter Sensitivity Analysis}
\label{app:hyperparameter}
We investigated the impact of the most significant parameters of Leddam: dimension of layer ($dim$), number of network layers ($nl$), dropout ratio ($dr$), and size of decomposition kernel ($k$). The default settings were: $dr=0.0$, $dim=512$, $nl=2$, $k=25$. Based on Table~\ref{t:dr}–~\ref{t:ks}, we can easily conclude that \textbf{Leddam} is insensitive to dropout ratio and kernel\_size, while more sensitive to the number of network layers and the dimension of layers. For the kernel size, we argue after our initialization with a Gaussian distribution of $0$ mean and $1$ variance, the weight of the central element is the largest. As moving away from the center, the weights gradually decrease. Therefore, when the kernel size increases, the weights of the farther points become smaller, thus not significantly affecting the final decomposition result. 

\begin{table}[t]
\centering
\caption{Impact of dropout ratio ($dr$). $dr \in \{0.0, 0.1, 0.2, 0.5\}$}
\label{t:dr}
\resizebox{0.5\textwidth}{!}{
\begin{tabular}{c|cc|cc}
\toprule
Dataset & \multicolumn{2}{c}{ETTm1} & \multicolumn{2}{c}{ETTm2} \\
\midrule
$dr$/Metric & MSE & MAE & MSE & MAE \\
\midrule
0.0 & 0.469 & 0.447 & 0.406 & 0.400 \\
0.1 & 0.468 & 0.446 & 0.404 & 0.401 \\
0.2 & 0.476 & 0.452 & 0.406 & 0.401 \\
0.5 & 0.474 & 0.452 & 0.409 & 0.404 \\
\bottomrule
\end{tabular}
}
\end{table}

\begin{table}[t]
\centering
\caption{Impact of dimension of layer ($dim$). $dim \in \{128, 256, 512, 1024\}$}
\label{t:dim}
\resizebox{0.5\textwidth}{!}{
\begin{tabular}{c|cc|cc}
\toprule
Dataset & \multicolumn{2}{c}{ETTm1} & \multicolumn{2}{c}{ETTm2} \\
\midrule
$dim$/Metric & MSE & MAE & MSE & MAE \\
\midrule
128 & 0.486 & 0.454 & 0.399 & 0.397 \\
256 & 0.473 & 0.449 & 0.403 & 0.399 \\
512 & 0.469 & 0.447 & 0.406 & 0.400 \\
1024 & 0.492 & 0.459 & 0.402 & 0.397 \\
\bottomrule
\end{tabular}
}
\end{table}

\begin{table*}[t]
\centering
\caption{Impact of number of network layers ($nl$). $nl \in \{1, 2, 3, 4\}$}
\label{t:nl}
\resizebox{0.5\textwidth}{!}{
\begin{tabular}{c|cc|cc}
\toprule
Dataset & \multicolumn{2}{c}{ETTm1} & \multicolumn{2}{c}{ETTm2} \\
\midrule
$nl$/Metric & MSE & MAE & MSE & MAE \\
\midrule
1 & 0.474 & 0.447 & 0.398 & 0.398 \\
2 & 0.469 & 0.447 & 0.406 & 0.400 \\
3 & 0.482 & 0.456 & 0.427 & 0.413 \\
4 & 0.485 & 0.456 & 0.407 & 0.401 \\
\bottomrule
\end{tabular}
}
\end{table*}

\begin{table*}[t]
\centering
\caption{Impact of kernel\_size ($k$). $k \in \{15, 25, 55, 75, 105\}$}
\label{t:ks}
\resizebox{0.5\textwidth}{!}{
\begin{tabular}{c|cc|cc}
\toprule
Dataset & \multicolumn{2}{c}{ETTm1} & \multicolumn{2}{c}{ETTm2} \\
\midrule
$k$/Metric & MSE & MAE & MSE & MAE \\
\midrule
15 & 0.469 & 0.447 & 0.406 & 0.400 \\
25 & 0.469 & 0.447 & 0.406 & 0.400 \\
55 & 0.469 & 0.448 & 0.407 & 0.401 \\
75 & 0.470 & 0.446 & 0.406 & 0.401 \\
105 & 0.469 & 0.447 & 0.406 & 0.400 \\
\bottomrule
\end{tabular}
}
\end{table*}

\section{Full Forecasting Results}
\label{app:full results}
The full multivariate forecasting results are provided in the following section due to the space limitation of the main text. Table~\ref{t:full_forecasting_results} contains the detailed results of eight baselines and our Leddam on eight well-acknowledged forecasting benchmarks.
\begin{table*}[t]
   \vspace{-8mm}
    \caption{Multivariate long-term forecasting result comparison. We use prediction lengths $F \in \{96, 192, 336, 720\}$, and input length $T = 96$. 
    The best results are in \textbf{bold} and the second bests are \underline{underlined}. }\label{t:full_forecasting_results}
    \vskip 0.05in
    \centering
    \setlength{\tabcolsep}{0.75pt}
    \resizebox{0.90\textwidth}{!}{
    \begin{threeparttable}
    \begin{small}
    \renewcommand{\multirowsetup}{\centering}
    \setlength{\tabcolsep}{1pt}
    \begin{tabular}{c|c|cc|cc|cc|cc|cc|cc|cc|cc|cc}
 \toprule
 \multicolumn{2}{c}{Model} & 
 \multicolumn{2}{c}{\rotatebox{0}{\scalebox{0.8}{\textbf{Leddam}}}} &\multicolumn{2}{c}{\rotatebox{0}{\scalebox{0.8}{iTransformer}}} &\multicolumn{2}{c}{\rotatebox{0}{\scalebox{0.8}{TimesNet}}} &\multicolumn{2}{c}{\rotatebox{0}{\scalebox{0.8}{MICN}}} &\multicolumn{2}{c}{\rotatebox{0}{\scalebox{0.8}{DLinear}}} &\multicolumn{2}{c}{\rotatebox{0}{\scalebox{0.8}{PatchTST}}} &\multicolumn{2}{c}{\rotatebox{0}{\scalebox{0.8}{Crossformer}}} &\multicolumn{2}{c}{\rotatebox{0}{\scalebox{0.8}{TiDE}}} &\multicolumn{2}{c}{\rotatebox{0}{\scalebox{0.8}{SCINet}}} \\
 
 \multicolumn{2}{c}{ } & \multicolumn{2}{c}{\scalebox{0.76}{(\textbf{Ours})}} & 
 \multicolumn{2}{c}{\scalebox{0.76}{\citeyearpar{LiuiTransformer}}} & 
 \multicolumn{2}{c}{\scalebox{0.76}{\citeyearpar{wu2022timesnet}}} & 
 \multicolumn{2}{c}{\scalebox{0.76}{\citeyearpar{wang2023micn}}} & 
 \multicolumn{2}{c}{\scalebox{0.76}{\citeyearpar{zeng2023dlinear}}} & 
 \multicolumn{2}{c}{\scalebox{0.76}{\citeyearpar{Nie2022patchtst}}} & 
 \multicolumn{2}{c}{\scalebox{0.76}{\citeyearpar{zhang2023crossformer}}} & 
 \multicolumn{2}{c}{\scalebox{0.76}{\citeyearpar{Das2023TiDE}}} & 
 \multicolumn{2}{c}{\scalebox{0.76}{\citeyearpar{Liu2021scinet}}}  \\
 
 \cmidrule(lr){3-4}\cmidrule(lr){5-6}\cmidrule(lr){7-8}\cmidrule(lr){9-10}\cmidrule(lr){11-12}\cmidrule(lr){13-14}\cmidrule(lr){15-16}\cmidrule(lr){17-18}\cmidrule(lr){19-20}
 \multicolumn{2}{c}{Metric} & \scalebox{0.78}{MSE} & \scalebox{0.78}{MAE} & \scalebox{0.78}{MSE} & \scalebox{0.78}{MAE} & \scalebox{0.78}{MSE} & \scalebox{0.78}{MAE} & \scalebox{0.78}{MSE} & \scalebox{0.78}{MAE} & \scalebox{0.78}{MSE} & \scalebox{0.78}{MAE} & \scalebox{0.78}{MSE} & \scalebox{0.78}{MAE} & \scalebox{0.78}{MSE} & \scalebox{0.78}{MAE} & \scalebox{0.78}{MSE} & \scalebox{0.78}{MAE} & \scalebox{0.78}{MSE} & \scalebox{0.78}{MAE} \\
 \toprule
 
 \multirow{5}{*}{\rotatebox{90}{\scalebox{0.95}{ETTh1}}} 
& \scalebox{0.78}{96} & \textcolor{red}{\textbf{\scalebox{0.78}{0.377}}} & \textcolor{red}{\textbf{\scalebox{0.78}{0.394}}} & \scalebox{0.78}{0.386} & \scalebox{0.78}{0.405} & \scalebox{0.78}{0.384} & \scalebox{0.78}{0.402} & \scalebox{0.78}{0.421} & \scalebox{0.78}{0.431} & \scalebox{0.78}{0.386} & \scalebox{0.78}{0.400} & \scalebox{0.78}{0.378} & \textcolor{blue}{\underline{\scalebox{0.78}{0.396}}} & \scalebox{0.78}{0.420} & \scalebox{0.78}{0.439} & \textcolor{blue}{\underline{\scalebox{0.78}{0.377}}} & \scalebox{0.78}{0.397} & \scalebox{0.78}{0.404} & \scalebox{0.78}{0.415}\\
& \scalebox{0.78}{192} & \textcolor{red}{\textbf{\scalebox{0.78}{0.424}}} & \textcolor{red}{\textbf{\scalebox{0.78}{0.422}}} & \scalebox{0.78}{0.441} & \scalebox{0.78}{0.436} & \scalebox{0.78}{0.436} & \scalebox{0.78}{0.429} & \scalebox{0.78}{0.474} & \scalebox{0.78}{0.487} & \scalebox{0.78}{0.437} & \scalebox{0.78}{0.432} & \textcolor{blue}{\underline{\scalebox{0.78}{0.424}}} & \textcolor{blue}{\underline{\scalebox{0.78}{0.425}}} & \scalebox{0.78}{0.541} & \scalebox{0.78}{0.520} & \scalebox{0.78}{0.425} & \scalebox{0.78}{0.431} & \scalebox{0.78}{0.456} & \scalebox{0.78}{0.445}\\
& \scalebox{0.78}{336} & \textcolor{red}{\textbf{\scalebox{0.78}{0.459}}} & \textcolor{red}{\textbf{\scalebox{0.78}{0.442}}} & \scalebox{0.78}{0.487} & \scalebox{0.78}{0.458} & \scalebox{0.78}{0.491} & \scalebox{0.78}{0.469} & \scalebox{0.78}{0.569} & \scalebox{0.78}{0.551} & \scalebox{0.78}{0.481} & \scalebox{0.78}{0.459} & \scalebox{0.78}{0.466} & \scalebox{0.78}{0.448} & \scalebox{0.78}{0.722} & \scalebox{0.78}{0.648} & \textcolor{blue}{\underline{\scalebox{0.78}{0.461}}} & \textcolor{blue}{\underline{\scalebox{0.78}{0.443}}} & \scalebox{0.78}{0.519} & \scalebox{0.78}{0.481}\\
& \scalebox{0.78}{720} & \textcolor{red}{\textbf{\scalebox{0.78}{0.463}}} & \textcolor{red}{\textbf{\scalebox{0.78}{0.459}}} & \scalebox{0.78}{0.503} & \scalebox{0.78}{0.491} & \scalebox{0.78}{0.521} & \scalebox{0.78}{0.500} & \scalebox{0.78}{0.770} & \scalebox{0.78}{0.672} & \scalebox{0.78}{0.519} & \scalebox{0.78}{0.516} & \scalebox{0.78}{0.529} & \scalebox{0.78}{0.500} & \scalebox{0.78}{0.814} & \scalebox{0.78}{0.692} & \textcolor{blue}{\underline{\scalebox{0.78}{0.471}}} & \textcolor{blue}{\underline{\scalebox{0.78}{0.478}}} & \scalebox{0.78}{0.564} & \scalebox{0.78}{0.528}\\ 
\cmidrule(lr){2-20}
& \scalebox{0.78}{Avg} & \textcolor{red}{\textbf{\scalebox{0.78}{0.431}}} & \textcolor{red}{\textbf{\scalebox{0.78}{0.429}}} & \scalebox{0.78}{0.454} & \scalebox{0.78}{0.447} & \scalebox{0.78}{0.458} & \scalebox{0.78}{0.450} & \scalebox{0.78}{0.561} & \scalebox{0.78}{0.535} & \scalebox{0.78}{0.456} & \scalebox{0.78}{0.452} & \scalebox{0.78}{0.449} & \scalebox{0.78}{0.442} & \scalebox{0.78}{0.624} & \scalebox{0.78}{0.575} & \textcolor{blue}{\underline{\scalebox{0.78}{0.434}}} & \textcolor{blue}{\underline{\scalebox{0.78}{0.437}}} & \scalebox{0.78}{0.486} & \scalebox{0.78}{0.467}\\ \midrule
\multirow{5}{*}{\rotatebox{90}{\scalebox{0.95}{ETTh2}}} 
& \scalebox{0.78}{96} & \textcolor{red}{\textbf{\scalebox{0.78}{0.292}}} & \textcolor{red}{\textbf{\scalebox{0.78}{0.343}}} & \textcolor{blue}{\underline{\scalebox{0.78}{0.297}}} & \scalebox{0.78}{0.349} & \scalebox{0.78}{0.340} & \scalebox{0.78}{0.374} & \scalebox{0.78}{0.299} & \scalebox{0.78}{0.364} & \scalebox{0.78}{0.333} & \scalebox{0.78}{0.387} & \scalebox{0.78}{0.302} & \textcolor{blue}{\underline{\scalebox{0.78}{0.348}}} & \scalebox{0.78}{0.745} & \scalebox{0.78}{0.584} & \scalebox{0.78}{0.400} & \scalebox{0.78}{0.440} & \scalebox{0.78}{0.707} & \scalebox{0.78}{0.621}\\
& \scalebox{0.78}{192} & \textcolor{red}{\textbf{\scalebox{0.78}{0.367}}} & \textcolor{red}{\textbf{\scalebox{0.78}{0.389}}} & \textcolor{blue}{\underline{\scalebox{0.78}{0.380}}} & \textcolor{blue}{\underline{\scalebox{0.78}{0.400}}} & \scalebox{0.78}{0.402} & \scalebox{0.78}{0.414} & \scalebox{0.78}{0.441} & \scalebox{0.78}{0.454} & \scalebox{0.78}{0.477} & \scalebox{0.78}{0.476} & \scalebox{0.78}{0.388} & \scalebox{0.78}{0.400} & \scalebox{0.78}{0.877} & \scalebox{0.78}{0.656} & \scalebox{0.78}{0.528} & \scalebox{0.78}{0.509} & \scalebox{0.78}{0.860} & \scalebox{0.78}{0.689}\\
& \scalebox{0.78}{336} & \textcolor{red}{\textbf{\scalebox{0.78}{0.412}}} & \textcolor{red}{\textbf{\scalebox{0.78}{0.424}}} & \scalebox{0.78}{0.428} & \textcolor{blue}{\underline{\scalebox{0.78}{0.432}}} & \scalebox{0.78}{0.452} & \scalebox{0.78}{0.452} & \scalebox{0.78}{0.654} & \scalebox{0.78}{0.567} & \scalebox{0.78}{0.594} & \scalebox{0.78}{0.541} & \textcolor{blue}{\underline{\scalebox{0.78}{0.426}}} & \scalebox{0.78}{0.433} & \scalebox{0.78}{1.043} & \scalebox{0.78}{0.731} & \scalebox{0.78}{0.643} & \scalebox{0.78}{0.571} & \scalebox{0.78}{1.000} & \scalebox{0.78}{0.744}\\
& \scalebox{0.78}{720} & \textcolor{red}{\textbf{\scalebox{0.78}{0.419}}} & \textcolor{red}{\textbf{\scalebox{0.78}{0.438}}} & \textcolor{blue}{\underline{\scalebox{0.78}{0.427}}} & \textcolor{blue}{\underline{\scalebox{0.78}{0.445}}} & \scalebox{0.78}{0.462} & \scalebox{0.78}{0.468} & \scalebox{0.78}{0.956} & \scalebox{0.78}{0.716} & \scalebox{0.78}{0.831} & \scalebox{0.78}{0.657} & \scalebox{0.78}{0.431} & \scalebox{0.78}{0.446} & \scalebox{0.78}{1.104} & \scalebox{0.78}{0.763} & \scalebox{0.78}{0.874} & \scalebox{0.78}{0.679} & \scalebox{0.78}{1.249} & \scalebox{0.78}{0.838}\\ 
\cmidrule(lr){2-20}
& \scalebox{0.78}{Avg} & \textcolor{red}{\textbf{\scalebox{0.78}{0.373}}} & \textcolor{red}{\textbf{\scalebox{0.78}{0.399}}} & \textcolor{blue}{\underline{\scalebox{0.78}{0.383}}} & \textcolor{blue}{\underline{\scalebox{0.78}{0.407}}} & \scalebox{0.78}{0.414} & \scalebox{0.78}{0.427} & \scalebox{0.78}{0.587} & \scalebox{0.78}{0.525} & \scalebox{0.78}{0.559} & \scalebox{0.78}{0.515} & \scalebox{0.78}{0.387} & \scalebox{0.78}{0.407} & \scalebox{0.78}{0.942} & \scalebox{0.78}{0.684} & \scalebox{0.78}{0.611} & \scalebox{0.78}{0.550} & \scalebox{0.78}{0.954} & \scalebox{0.78}{0.723}\\  \midrule
\multirow{5}{*}{\rotatebox{90}{\scalebox{0.95}{ETTm1}}} 
& \scalebox{0.78}{96} & \textcolor{red}{\textbf{\scalebox{0.78}{0.319}}} & \textcolor{red}{\textbf{\scalebox{0.78}{0.359}}} & \scalebox{0.78}{0.334} & \scalebox{0.78}{0.368} & \scalebox{0.78}{0.338} & \scalebox{0.78}{0.375} & \scalebox{0.78}{0.324} & \scalebox{0.78}{0.375} & \scalebox{0.78}{0.345} & \scalebox{0.78}{0.372} & \textcolor{blue}{\underline{\scalebox{0.78}{0.322}}} & \textcolor{blue}{\underline{\scalebox{0.78}{0.362}}} & \scalebox{0.78}{0.360} & \scalebox{0.78}{0.401} & \scalebox{0.78}{0.347} & \scalebox{0.78}{0.384} & \scalebox{0.78}{0.350} & \scalebox{0.78}{0.385} \\
& \scalebox{0.78}{192} & \textcolor{blue}{\underline{\scalebox{0.78}{0.369}}} & \textcolor{red}{\textbf{\scalebox{0.78}{0.383}}} & \scalebox{0.78}{0.377} & \scalebox{0.78}{0.391} & \scalebox{0.78}{0.374} & \scalebox{0.78}{0.387} & \scalebox{0.78}{0.366} & \scalebox{0.78}{0.402} & \scalebox{0.78}{0.380} & \scalebox{0.78}{0.389} & \textcolor{red}{\textbf{\scalebox{0.78}{0.366}}} & \textcolor{blue}{\underline{\scalebox{0.78}{0.387}}} & \scalebox{0.78}{0.403} & \scalebox{0.78}{0.440} & \scalebox{0.78}{0.397} & \scalebox{0.78}{0.409} & \scalebox{0.78}{0.382} & \scalebox{0.78}{0.400} \\
& \scalebox{0.78}{336} & \textcolor{red}{\textbf{\scalebox{0.78}{0.394}}} & \textcolor{red}{\textbf{\scalebox{0.78}{0.402}}} & \scalebox{0.78}{0.426} & \scalebox{0.78}{0.420} & \scalebox{0.78}{0.410} & \scalebox{0.78}{0.411} & \scalebox{0.78}{0.408} & \scalebox{0.78}{0.426} & \scalebox{0.78}{0.413} & \scalebox{0.78}{0.413} & \textcolor{blue}{\underline{\scalebox{0.78}{0.396}}} & \textcolor{blue}{\underline{\scalebox{0.78}{0.404}}} & \scalebox{0.78}{0.543} & \scalebox{0.78}{0.528} & \scalebox{0.78}{0.417} & \scalebox{0.78}{0.430} & \scalebox{0.78}{0.419} & \scalebox{0.78}{0.425} \\
& \scalebox{0.78}{720} & \textcolor{red}{\textbf{\scalebox{0.78}{0.460}}} & \textcolor{red}{\textbf{\scalebox{0.78}{0.442}}} & \scalebox{0.78}{0.491} & \scalebox{0.78}{0.459} & \scalebox{0.78}{0.478} & \scalebox{0.78}{0.450} & \scalebox{0.78}{0.481} & \scalebox{0.78}{0.476} & \scalebox{0.78}{0.474} & \scalebox{0.78}{0.453} & \textcolor{blue}{\underline{\scalebox{0.78}{0.464}}} & \textcolor{blue}{\underline{\scalebox{0.78}{0.446}}} & \scalebox{0.78}{0.744} & \scalebox{0.78}{0.666} & \scalebox{0.78}{0.472} & \scalebox{0.78}{0.485} & \scalebox{0.78}{0.494} & \scalebox{0.78}{0.463} \\ 
\cmidrule(lr){2-20}
& \scalebox{0.78}{Avg} & \textcolor{red}{\textbf{\scalebox{0.78}{0.386}}} & \textcolor{red}{\textbf{\scalebox{0.78}{0.397}}} & \scalebox{0.78}{0.407} & \scalebox{0.78}{0.410} & \scalebox{0.78}{0.400} & \scalebox{0.78}{0.406} & \scalebox{0.78}{0.392} & \scalebox{0.78}{0.414} & \scalebox{0.78}{0.403} & \scalebox{0.78}{0.407} & \textcolor{blue}{\underline{\scalebox{0.78}{0.387}}} & \textcolor{blue}{\underline{\scalebox{0.78}{0.400}}} & \scalebox{0.78}{0.513} & \scalebox{0.78}{0.509} & \scalebox{0.78}{0.403} & \scalebox{0.78}{0.427} & \scalebox{0.78}{0.411} & \scalebox{0.78}{0.418} \\ \midrule
\multirow{5}{*}{\rotatebox{90}{\scalebox{0.95}{ETTm2}}} 
& \scalebox{0.78}{96} & \textcolor{red}{\textbf{\scalebox{0.78}{0.176}}} & \textcolor{red}{\textbf{\scalebox{0.78}{0.257}}} & \scalebox{0.78}{0.180} & \scalebox{0.78}{0.264} & \scalebox{0.78}{0.187} & \scalebox{0.78}{0.267} & \scalebox{0.78}{0.179} & \scalebox{0.78}{0.275} & \scalebox{0.78}{0.193} & \scalebox{0.78}{0.292} & \textcolor{blue}{\underline{\scalebox{0.78}{0.178}}} & \textcolor{blue}{\underline{\scalebox{0.78}{0.260}}} & \scalebox{0.78}{0.259} & \scalebox{0.78}{0.349} & \scalebox{0.78}{0.192} & \scalebox{0.78}{0.274} & \scalebox{0.78}{0.201} & \scalebox{0.78}{0.280} \\
& \scalebox{0.78}{192} & \textcolor{blue}{\underline{\scalebox{0.78}{0.243}}} & \textcolor{blue}{\underline{\scalebox{0.78}{0.303}}} & \scalebox{0.78}{0.250} & \scalebox{0.78}{0.309} & \scalebox{0.78}{0.249} & \scalebox{0.78}{0.309} & \scalebox{0.78}{0.307} & \scalebox{0.78}{0.376} & \scalebox{0.78}{0.284} & \scalebox{0.78}{0.362} & \textcolor{red}{\textbf{\scalebox{0.78}{0.242}}} & \textcolor{red}{\textbf{\scalebox{0.78}{0.301}}} & \scalebox{0.78}{0.543} & \scalebox{0.78}{0.551} & \scalebox{0.78}{0.253} & \scalebox{0.78}{0.313} & \scalebox{0.78}{0.283} & \scalebox{0.78}{0.331} \\
& \scalebox{0.78}{336} & \textcolor{red}{\textbf{\scalebox{0.78}{0.303}}} & \textcolor{red}{\textbf{\scalebox{0.78}{0.341}}} & \scalebox{0.78}{0.311} & \scalebox{0.78}{0.348} & \scalebox{0.78}{0.321} & \scalebox{0.78}{0.351} & \scalebox{0.78}{0.325} & \scalebox{0.78}{0.388} & \scalebox{0.78}{0.369} & \scalebox{0.78}{0.427} & \textcolor{blue}{\underline{\scalebox{0.78}{0.304}}} & \textcolor{blue}{\underline{\scalebox{0.78}{0.344}}} & \scalebox{0.78}{1.038} & \scalebox{0.78}{0.715} & \scalebox{0.78}{0.315} & \scalebox{0.78}{0.352} & \scalebox{0.78}{0.318} & \scalebox{0.78}{0.352} \\
& \scalebox{0.78}{720} & \textcolor{red}{\textbf{\scalebox{0.78}{0.400}}} & \textcolor{red}{\textbf{\scalebox{0.78}{0.398}}} & \scalebox{0.78}{0.412} & \scalebox{0.78}{0.407} & \scalebox{0.78}{0.408} & \scalebox{0.78}{0.403} & \scalebox{0.78}{0.502} & \scalebox{0.78}{0.490} & \scalebox{0.78}{0.554} & \scalebox{0.78}{0.522} & \textcolor{blue}{\underline{\scalebox{0.78}{0.410}}} & \textcolor{blue}{\underline{\scalebox{0.78}{0.404}}} & \scalebox{0.78}{6.037} & \scalebox{0.78}{1.693} & \scalebox{0.78}{0.413} & \scalebox{0.78}{0.406} & \scalebox{0.78}{0.439} & \scalebox{0.78}{0.423} \\ 
\cmidrule(lr){2-20}
& \scalebox{0.78}{Avg} & \textcolor{red}{\textbf{\scalebox{0.78}{0.281}}} & \textcolor{red}{\textbf{\scalebox{0.78}{0.325}}} & \scalebox{0.78}{0.288} & \scalebox{0.78}{0.332} & \scalebox{0.78}{0.291} & \scalebox{0.78}{0.333} & \scalebox{0.78}{0.328} & \scalebox{0.78}{0.382} & \scalebox{0.78}{0.350} & \scalebox{0.78}{0.401} & \textcolor{blue}{\underline{\scalebox{0.78}{0.283}}} & \textcolor{blue}{\underline{\scalebox{0.78}{0.327}}} & \scalebox{0.78}{1.219} & \scalebox{0.78}{0.827} & \scalebox{0.78}{0.293} & \scalebox{0.78}{0.336} & \scalebox{0.78}{0.310} & \scalebox{0.78}{0.347} \\ \midrule
\multirow{5}{*}{\rotatebox{90}{\scalebox{0.95}{Electricity}}} 
& \scalebox{0.78}{96} & \textcolor{red}{\textbf{\scalebox{0.78}{0.141}}} & \textcolor{red}{\textbf{\scalebox{0.78}{0.235}}} & \textcolor{blue}{\underline{\scalebox{0.78}{0.148}}} & \textcolor{blue}{\underline{\scalebox{0.78}{0.240}}} & \scalebox{0.78}{0.168} & \scalebox{0.78}{0.272} & \scalebox{0.78}{0.164} & \scalebox{0.78}{0.269} & \scalebox{0.78}{0.197} & \scalebox{0.78}{0.282} & \scalebox{0.78}{0.195} & \scalebox{0.78}{0.285} & \scalebox{0.78}{0.219} & \scalebox{0.78}{0.314} & \scalebox{0.78}{0.237} & \scalebox{0.78}{0.329} & \scalebox{0.78}{0.247} & \scalebox{0.78}{0.345} \\
& \scalebox{0.78}{192} & \textcolor{red}{\textbf{\scalebox{0.78}{0.159}}} & \textcolor{red}{\textbf{\scalebox{0.78}{0.252}}} & \textcolor{blue}{\underline{\scalebox{0.78}{0.162}}} & \textcolor{blue}{\underline{\scalebox{0.78}{0.253}}} & \scalebox{0.78}{0.184} & \scalebox{0.78}{0.289} & \scalebox{0.78}{0.177} & \scalebox{0.78}{0.285} & \scalebox{0.78}{0.196} & \scalebox{0.78}{0.285} & \scalebox{0.78}{0.199} & \scalebox{0.78}{0.289} & \scalebox{0.78}{0.231} & \scalebox{0.78}{0.322} & \scalebox{0.78}{0.236} & \scalebox{0.78}{0.330} & \scalebox{0.78}{0.257} & \scalebox{0.78}{0.355} \\
& \scalebox{0.78}{336} & \textcolor{red}{\textbf{\scalebox{0.78}{0.173}}} & \textcolor{red}{\textbf{\scalebox{0.78}{0.268}}} & \textcolor{blue}{\underline{\scalebox{0.78}{0.178}}} & \textcolor{blue}{\underline{\scalebox{0.78}{0.269}}} & \scalebox{0.78}{0.198} & \scalebox{0.78}{0.300} & \scalebox{0.78}{0.193} & \scalebox{0.78}{0.304} & \scalebox{0.78}{0.209} & \scalebox{0.78}{0.301} & \scalebox{0.78}{0.215} & \scalebox{0.78}{0.305} & \scalebox{0.78}{0.246} & \scalebox{0.78}{0.337} & \scalebox{0.78}{0.249} & \scalebox{0.78}{0.344} & \scalebox{0.78}{0.269} & \scalebox{0.78}{0.369} \\
& \scalebox{0.78}{720} & \textcolor{red}{\textbf{\scalebox{0.78}{0.201}}} & \textcolor{red}{\textbf{\scalebox{0.78}{0.295}}} & \scalebox{0.78}{0.225} & \textcolor{blue}{\underline{\scalebox{0.78}{0.317}}} & \scalebox{0.78}{0.220} & \scalebox{0.78}{0.320} & \textcolor{blue}{\underline{\scalebox{0.78}{0.212}}} & \scalebox{0.78}{0.321} & \scalebox{0.78}{0.245} & \scalebox{0.78}{0.333} & \scalebox{0.78}{0.256} & \scalebox{0.78}{0.337} & \scalebox{0.78}{0.280} & \scalebox{0.78}{0.363} & \scalebox{0.78}{0.284} & \scalebox{0.78}{0.373} & \scalebox{0.78}{0.299} & \scalebox{0.78}{0.390} \\ 
\cmidrule(lr){2-20}
& \scalebox{0.78}{Avg} & \textcolor{red}{\textbf{\scalebox{0.78}{0.169}}} & \textcolor{red}{\textbf{\scalebox{0.78}{0.263}}} & \textcolor{blue}{\underline{\scalebox{0.78}{0.178}}} & \textcolor{blue}{\underline{\scalebox{0.78}{0.270}}} & \scalebox{0.78}{0.192} & \scalebox{0.78}{0.295} & \scalebox{0.78}{0.187} & \scalebox{0.78}{0.295} & \scalebox{0.78}{0.212} & \scalebox{0.78}{0.300} & \scalebox{0.78}{0.216} & \scalebox{0.78}{0.304} & \scalebox{0.78}{0.244} & \scalebox{0.78}{0.334} & \scalebox{0.78}{0.251} & \scalebox{0.78}{0.344} & \scalebox{0.78}{0.268} & \scalebox{0.78}{0.365} \\ \midrule
\multirow{5}{*}{\rotatebox{90}{\scalebox{0.95}{Solar\_Energy}}} 
& \scalebox{0.78}{96} & \textcolor{red}{\textbf{\scalebox{0.78}{0.197}}} & \textcolor{blue}{\underline{\scalebox{0.78}{0.241}}} & \textcolor{blue}{\underline{\scalebox{0.78}{0.203}}} & \textcolor{red}{\textbf{\scalebox{0.78}{0.237}}} & \scalebox{0.78}{0.250} & \scalebox{0.78}{0.292} & \scalebox{0.78}{0.222} & \scalebox{0.78}{0.310} & \scalebox{0.78}{0.290} & \scalebox{0.78}{0.378} & \scalebox{0.78}{0.234} & \scalebox{0.78}{0.286} & \scalebox{0.78}{0.310} & \scalebox{0.78}{0.331} & \scalebox{0.78}{0.312} & \scalebox{0.78}{0.399} & \scalebox{0.78}{0.237} & \scalebox{0.78}{0.344} \\
& \scalebox{0.78}{192} & \textcolor{red}{\textbf{\scalebox{0.78}{0.231}}} & \textcolor{blue}{\underline{\scalebox{0.78}{0.264}}} & \textcolor{blue}{\underline{\scalebox{0.78}{0.233}}} & \textcolor{red}{\textbf{\scalebox{0.78}{0.261}}} & \scalebox{0.78}{0.296} & \scalebox{0.78}{0.318} & \scalebox{0.78}{0.277} & \scalebox{0.78}{0.343} & \scalebox{0.78}{0.320} & \scalebox{0.78}{0.398} & \scalebox{0.78}{0.267} & \scalebox{0.78}{0.310} & \scalebox{0.78}{0.734} & \scalebox{0.78}{0.725} & \scalebox{0.78}{0.339} & \scalebox{0.78}{0.416} & \scalebox{0.78}{0.280} & \scalebox{0.78}{0.380} \\
& \scalebox{0.78}{336} & \textcolor{red}{\textbf{\scalebox{0.78}{0.241}}} & \textcolor{red}{\textbf{\scalebox{0.78}{0.268}}} & \textcolor{blue}{\underline{\scalebox{0.78}{0.248}}} & \textcolor{blue}{\underline{\scalebox{0.78}{0.273}}} & \scalebox{0.78}{0.319} & \scalebox{0.78}{0.330} & \scalebox{0.78}{0.297} & \scalebox{0.78}{0.386} & \scalebox{0.78}{0.353} & \scalebox{0.78}{0.415} & \scalebox{0.78}{0.290} & \scalebox{0.78}{0.315} & \scalebox{0.78}{0.750} & \scalebox{0.78}{0.735} & \scalebox{0.78}{0.368} & \scalebox{0.78}{0.430} & \scalebox{0.78}{0.304} & \scalebox{0.78}{0.389} \\
& \scalebox{0.78}{720} & \textcolor{blue}{\underline{\scalebox{0.78}{0.250}}} & \textcolor{blue}{\underline{\scalebox{0.78}{0.281}}} & \textcolor{red}{\textbf{\scalebox{0.78}{0.249}}} & \textcolor{red}{\textbf{\scalebox{0.78}{0.275}}} & \scalebox{0.78}{0.338} & \scalebox{0.78}{0.337} & \scalebox{0.78}{0.390} & \scalebox{0.78}{0.445} & \scalebox{0.78}{0.356} & \scalebox{0.78}{0.413} & \scalebox{0.78}{0.289} & \scalebox{0.78}{0.317} & \scalebox{0.78}{0.769} & \scalebox{0.78}{0.765} & \scalebox{0.78}{0.370} & \scalebox{0.78}{0.425} & \scalebox{0.78}{0.308} & \scalebox{0.78}{0.388} \\ 
\cmidrule(lr){2-20}
& \scalebox{0.78}{Avg} & \textcolor{red}{\textbf{\scalebox{0.78}{0.230}}} & \textcolor{blue}{\underline{\scalebox{0.78}{0.264}}} & \textcolor{blue}{\underline{\scalebox{0.78}{0.233}}} & \textcolor{red}{\textbf{\scalebox{0.78}{0.262}}} & \scalebox{0.78}{0.301} & \scalebox{0.78}{0.319} & \scalebox{0.78}{0.296} & \scalebox{0.78}{0.371} & \scalebox{0.78}{0.330} & \scalebox{0.78}{0.401} & \scalebox{0.78}{0.270} & \scalebox{0.78}{0.307} & \scalebox{0.78}{0.641} & \scalebox{0.78}{0.639} & \scalebox{0.78}{0.347} & \scalebox{0.78}{0.417} & \scalebox{0.78}{0.282} & \scalebox{0.78}{0.375} \\ \midrule
\multirow{5}{*}{\rotatebox{90}{\scalebox{0.95}{Traffic}}} 
& \scalebox{0.78}{96} & \textcolor{blue}{\underline{\scalebox{0.78}{0.426}}} & \textcolor{blue}{\underline{\scalebox{0.78}{0.276}}} & \textcolor{red}{\textbf{\scalebox{0.78}{0.395}}} & \textcolor{red}{\textbf{\scalebox{0.78}{0.268}}} & \scalebox{0.78}{0.593} & \scalebox{0.78}{0.321} & \scalebox{0.78}{0.519} & \scalebox{0.78}{0.309} & \scalebox{0.78}{0.650} & \scalebox{0.78}{0.396} & \scalebox{0.78}{0.544} & \scalebox{0.78}{0.359} & \scalebox{0.78}{0.522} & \scalebox{0.78}{0.290} & \scalebox{0.78}{0.805} & \scalebox{0.78}{0.493} & \scalebox{0.78}{0.788} & \scalebox{0.78}{0.499} \\
& \scalebox{0.78}{192} & \textcolor{blue}{\underline{\scalebox{0.78}{0.458}}} & \textcolor{blue}{\underline{\scalebox{0.78}{0.289}}} & \textcolor{red}{\textbf{\scalebox{0.78}{0.417}}} & \textcolor{red}{\textbf{\scalebox{0.78}{0.276}}} & \scalebox{0.78}{0.617} & \scalebox{0.78}{0.336} & \scalebox{0.78}{0.537} & \scalebox{0.78}{0.315} & \scalebox{0.78}{0.598} & \scalebox{0.78}{0.370} & \scalebox{0.78}{0.540} & \scalebox{0.78}{0.354} & \scalebox{0.78}{0.530} & \scalebox{0.78}{0.293} & \scalebox{0.78}{0.756} & \scalebox{0.78}{0.474} & \scalebox{0.78}{0.789} & \scalebox{0.78}{0.505} \\
& \scalebox{0.78}{336} & \textcolor{blue}{\underline{\scalebox{0.78}{0.486}}} & \textcolor{blue}{\underline{\scalebox{0.78}{0.297}}} & \textcolor{red}{\textbf{\scalebox{0.78}{0.433}}} & \textcolor{red}{\textbf{\scalebox{0.78}{0.283}}} & \scalebox{0.78}{0.629} & \scalebox{0.78}{0.336} & \scalebox{0.78}{0.534} & \scalebox{0.78}{0.313} & \scalebox{0.78}{0.605} & \scalebox{0.78}{0.373} & \scalebox{0.78}{0.551} & \scalebox{0.78}{0.358} & \scalebox{0.78}{0.558} & \scalebox{0.78}{0.305} & \scalebox{0.78}{0.762} & \scalebox{0.78}{0.477} & \scalebox{0.78}{0.797} & \scalebox{0.78}{0.508} \\
& \scalebox{0.78}{720} & \textcolor{blue}{\underline{\scalebox{0.78}{0.498}}} & \textcolor{blue}{\underline{\scalebox{0.78}{0.313}}} & \textcolor{red}{\textbf{\scalebox{0.78}{0.467}}} & \textcolor{red}{\textbf{\scalebox{0.78}{0.302}}} & \scalebox{0.78}{0.640} & \scalebox{0.78}{0.350} & \scalebox{0.78}{0.577} & \scalebox{0.78}{0.325} & \scalebox{0.78}{0.645} & \scalebox{0.78}{0.394} & \scalebox{0.78}{0.586} & \scalebox{0.78}{0.375} & \scalebox{0.78}{0.589} & \scalebox{0.78}{0.328} & \scalebox{0.78}{0.719} & \scalebox{0.78}{0.449} & \scalebox{0.78}{0.841} & \scalebox{0.78}{0.523} \\ 
\cmidrule(lr){2-20}
& \scalebox{0.78}{Avg} & \textcolor{blue}{\underline{\scalebox{0.78}{0.467}}} & \textcolor{blue}{\underline{\scalebox{0.78}{0.294}}} & \textcolor{red}{\textbf{\scalebox{0.78}{0.428}}} & \textcolor{red}{\textbf{\scalebox{0.78}{0.282}}} & \scalebox{0.78}{0.620} & \scalebox{0.78}{0.336} & \scalebox{0.78}{0.542} & \scalebox{0.78}{0.315} & \scalebox{0.78}{0.625} & \scalebox{0.78}{0.383} & \scalebox{0.78}{0.555} & \scalebox{0.78}{0.362} & \scalebox{0.78}{0.550} & \scalebox{0.78}{0.304} & \scalebox{0.78}{0.760} & \scalebox{0.78}{0.473} & \scalebox{0.78}{0.804} & \scalebox{0.78}{0.509} \\ \midrule
\multirow{5}{*}{\rotatebox{90}{\scalebox{0.95}{Weather}}} 
& \scalebox{0.78}{96} & \textcolor{red}{\textbf{\scalebox{0.78}{0.156}}} & \textcolor{red}{\textbf{\scalebox{0.78}{0.202}}} & \scalebox{0.78}{0.174} & \textcolor{blue}{\underline{\scalebox{0.78}{0.214}}} & \scalebox{0.78}{0.172} & \scalebox{0.78}{0.220} & \scalebox{0.78}{0.161} & \scalebox{0.78}{0.229} & \scalebox{0.78}{0.196} & \scalebox{0.78}{0.255} & \scalebox{0.78}{0.177} & \scalebox{0.78}{0.218} & \textcolor{blue}{\underline{\scalebox{0.78}{0.158}}} & \scalebox{0.78}{0.230} & \scalebox{0.78}{0.202} & \scalebox{0.78}{0.261} & \scalebox{0.78}{0.221} & \scalebox{0.78}{0.306} \\
& \scalebox{0.78}{192} & \textcolor{blue}{\underline{\scalebox{0.78}{0.207}}} & \textcolor{red}{\textbf{\scalebox{0.78}{0.250}}} & \scalebox{0.78}{0.221} & \textcolor{blue}{\underline{\scalebox{0.78}{0.254}}} & \scalebox{0.78}{0.219} & \scalebox{0.78}{0.261} & \scalebox{0.78}{0.220} & \scalebox{0.78}{0.281} & \scalebox{0.78}{0.237} & \scalebox{0.78}{0.296} & \scalebox{0.78}{0.225} & \scalebox{0.78}{0.259} & \textcolor{red}{\textbf{\scalebox{0.78}{0.206}}} & \scalebox{0.78}{0.277} & \scalebox{0.78}{0.242} & \scalebox{0.78}{0.298} & \scalebox{0.78}{0.261} & \scalebox{0.78}{0.340} \\
& \scalebox{0.78}{336} & \textcolor{red}{\textbf{\scalebox{0.78}{0.262}}} & \textcolor{red}{\textbf{\scalebox{0.78}{0.291}}} & \scalebox{0.78}{0.278} & \textcolor{blue}{\underline{\scalebox{0.78}{0.296}}} & \scalebox{0.78}{0.280} & \scalebox{0.78}{0.306} & \scalebox{0.78}{0.278} & \scalebox{0.78}{0.331} & \scalebox{0.78}{0.283} & \scalebox{0.78}{0.335} & \scalebox{0.78}{0.278} & \scalebox{0.78}{0.297} & \textcolor{blue}{\underline{\scalebox{0.78}{0.272}}} & \scalebox{0.78}{0.335} & \scalebox{0.78}{0.287} & \scalebox{0.78}{0.335} & \scalebox{0.78}{0.309} & \scalebox{0.78}{0.378} \\
& \scalebox{0.78}{720} & \textcolor{red}{\textbf{\scalebox{0.78}{0.343}}} & \textcolor{red}{\textbf{\scalebox{0.78}{0.343}}} & \scalebox{0.78}{0.358} & \scalebox{0.78}{0.349} & \scalebox{0.78}{0.365} & \scalebox{0.78}{0.359} & \scalebox{0.78}{0.311} & \scalebox{0.78}{0.356} & \scalebox{0.78}{0.345} & \scalebox{0.78}{0.381} & \textcolor{blue}{\underline{\scalebox{0.78}{0.354}}} & \textcolor{blue}{\underline{\scalebox{0.78}{0.348}}} & \scalebox{0.78}{0.398} & \scalebox{0.78}{0.418} & \scalebox{0.78}{0.351} & \scalebox{0.78}{0.386} & \scalebox{0.78}{0.377} & \scalebox{0.78}{0.427} \\
\cmidrule(lr){2-20}
& \scalebox{0.78}{Avg} & \textcolor{red}{\textbf{\scalebox{0.78}{0.242}}} & \textcolor{red}{\textbf{\scalebox{0.78}{0.272}}} & \textcolor{blue}{\underline{\scalebox{0.78}{0.258}}} & \textcolor{blue}{\underline{\scalebox{0.78}{0.279}}} & \scalebox{0.78}{0.259} & \scalebox{0.78}{0.287} & \scalebox{0.78}{0.243} & \scalebox{0.78}{0.299} & \scalebox{0.78}{0.265} & \scalebox{0.78}{0.317} & \scalebox{0.78}{0.259} & \scalebox{0.78}{0.281} & \scalebox{0.78}{0.259} & \scalebox{0.78}{0.315} & \scalebox{0.78}{0.271} & \scalebox{0.78}{0.320} & \scalebox{0.78}{0.292} & \scalebox{0.78}{0.363} \\ \midrule

& \scalebox{0.78}{$1^{st}$ Count} & \textcolor{red}{\textbf{\scalebox{0.78}{31}}} & \textcolor{red}{\textbf{\scalebox{0.78}{30}}} & \textcolor{blue}{\underline{\scalebox{0.78}{6}}} & \textcolor{blue}{\underline{\scalebox{0.78}{9}}} & \scalebox{0.78}{0} & \scalebox{0.78}{0} & \scalebox{0.78}{0} & \scalebox{0.78}{0} & \scalebox{0.78}{0} & \scalebox{0.78}{0} & \scalebox{0.78}{2} & \scalebox{0.78}{1} & \scalebox{0.78}{1} & \scalebox{0.78}{0} & \scalebox{0.78}{0} & \scalebox{0.78}{0} & \scalebox{0.78}{0} & \scalebox{0.78}{0} \\
             \bottomrule
          \end{tabular}
       \end{small}
    \end{threeparttable}}
     \vspace{-3mm}
 \end{table*}

\section{Influence of Input Length on Prediction Performance}
\label{app:inp}
\begin{figure*}[t]
   \centering
   \includegraphics[width=1.0\textwidth]{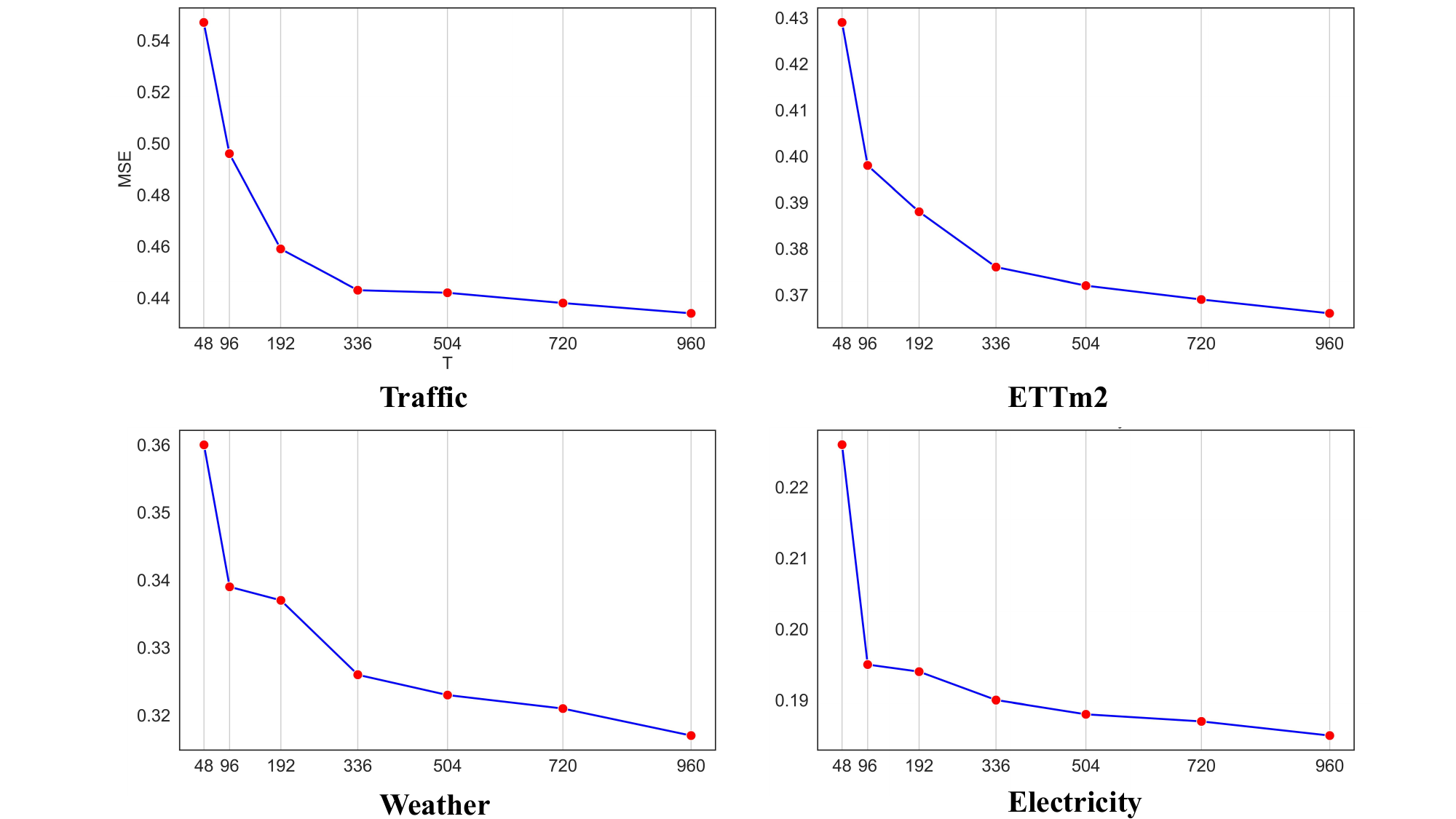}
   \caption{Forecasting performance (MSE) of Leddam with varying look-back windows on 4 datasets: ETTm2, Electricity, Traffic, and Weather. The look-back windows are selected to be $T \in \{48,96, 192, 336, 504, 720, 960\}$, and the prediction horizons are $F = 720$.}
   \label{figure:seq-length-compare}
\end{figure*}

In principle, extending the look-back window increases the receptive field, leading to a potential improvement in forecasting performance. A robust Time Series Forecasting (TSF) model equipped with a strong temporal relation extraction capability should yield improved results with larger look-back window sizes~\citep{zeng2023dlinear}. As demonstrated in Figure~\ref{figure:seq-length-compare}, Our Leddam model consistently and effectively diminishes MSE scores as the receptive field expands, affirming its capacity to leverage longer look-back windows and superior temporal relation extraction capabilities.

\end{document}